\documentclass[final]{cvpr}

\usepackage{times}
\usepackage{epsfig}
\usepackage{graphicx}
\usepackage{amsmath}
\usepackage{amssymb}

\usepackage{booktabs}
\usepackage{dsfont}
\usepackage[percent]{overpic}

\usepackage[pagebackref=true,breaklinks=true,colorlinks,bookmarks=false]{hyperref}

\newcommand{\denselist}{\itemsep 0pt\parsep=0pt\partopsep 0pt\vspace{-\topsep}}
\newcommand{\bitem}{\begin{itemize}\denselist}
\newcommand{\eitem}{\end{itemize}}
\newcommand{\modelname}{DANR}
\usepackage{subfigure}

\begin{document}
	
	\title{Data Augmentation for Object Detection via \\ Differentiable Neural Rendering}
	
	\author{Guanghan Ning\textsuperscript{1},\quad    Guang Chen\textsuperscript{1},\quad    Chaowei Tan\textsuperscript{1},\quad  Si Luo\textsuperscript{1},\quad   Liefeng Bo\textsuperscript{1},\quad   Heng Huang\textsuperscript{1, 2}\\
		\textsuperscript{1}JD Finance America Corporation\quad
		\textsuperscript{2}University of Pittsburgh\\
	}
	
	\maketitle

	\begin{abstract}
		It is challenging to train a robust object detector under the supervised learning setting when the annotated data are scarce. Thus, previous approaches tackling this problem are in two categories: semi-supervised learning models that interpolate labeled data from unlabeled data, and self-supervised learning approaches that exploit signals within unlabeled data via pretext tasks. To seamlessly integrate and enhance existing supervised object detection methods, in this work, we focus on addressing the data scarcity problem from a fundamental viewpoint without changing the supervised learning paradigm. 
		We propose a new offline data augmentation method for object detection, which semantically interpolates the training data with novel views.
		Specifically, our new system generates controllable views of training images based on differentiable neural rendering, together with corresponding bounding box annotations which involve no human intervention. 
		Firstly, we extract and project pixel-aligned image features into point clouds while estimating depth maps. We then re-project them with a target camera pose and render a novel-view 2d image. 
		Objects in the form of keypoints are marked in point clouds to recover annotations in new views.
		Our new method is fully compatible with online data augmentation methods, such as affine transform, image mixup, etc. Extensive experiments show that our method, as a cost-free tool to enrich images and labels, can significantly boost the performance of object detection systems with scarce training data. 
		Code is available at \url{https://github.com/Guanghan/DANR}.
	\end{abstract}
	
	\vspace{-5pt}
	\section{Introduction}
	Given a monocular RGB image, an object detection system aims to determine whether there are any instances of semantic objects from pre-defined categories and, if present, to locate the instance and return the confidence. 
	As a popular task, the object detection techniques have been widely deployed in real-world applications such as robotics and autonomous vehicles. However, in some applications, \emph{e.g.}, image content review, the data distribution is long-tailed, where labelled data of specific categories are naturally scarce. 
	Training data-driven neural networks under such circumstances is quite challenging in the supervised learning paradigm.
	First of all, training object detectors with scarce annotated data is inherently difficult. Detectors need to handle objects that may occur in stochastic regions of an image, thus requiring a significant amount of data.
	Secondly, traditional online data augmentation methods, including random crop, image mixup and dozens of others, provide additional data diversity to avoid over-fitting but do not provide unseen semantics with novel locations.
	An offline approach to enrich data diversity is to create synthetic datasets via generative networks.
	However, it is unnatural to simply stick foreground objects onto arbitrary background images (alpha compositing), which introduces artifacts that may undermine the model. Besides, it may still require some additional labeling, \emph{i.e.}, annotating foreground objects at pixel-level.
	
	\begin{figure}
		\centering
		\includegraphics[width=1.0\linewidth]{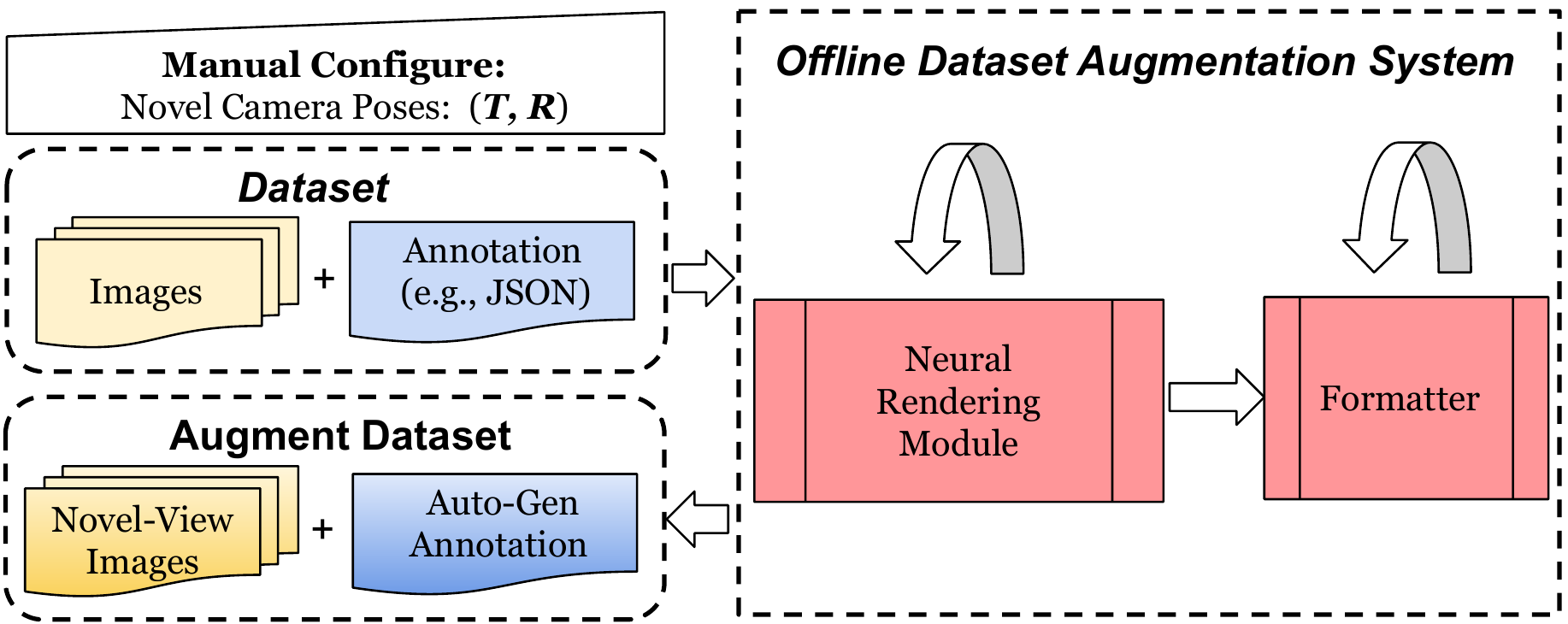}
		\caption{Overview of the proposed data augmentation system.} 
		\label{fig:overview}
		\vspace{-.2in}
	\end{figure}
	
	Several promising approaches were presented to tackle this problem, however, these methods are under different machine learning paradigms, \emph{e.g.}, via semi-supervised learning that interpolates labeled data from unlabeled data, or self-supervised learning that exploit signals within unlabeled data via pretext tasks.
	Without changing the supervised learning paradigm, we propose a new approach of offline data augmentation for object detection via novel-view synthesis based on the differentiable neural rendering. 
	The key motivation behind our approach is that novel views bring unseen yet realistic semantics and object locations, which are beneficial to a data-hungry object detector.
	Our system generates controllable views of training images given arbitrary novel camera poses, and produces corresponding bounding box annotations without human intervention. 
	Specifically, we extract and project pixel-aligned image features into latent space point clouds based on the estimated depth maps, and mark keypoints of the annotations. We then re-project them with a target camera pose to render the corresponding novel-view 2D image. Keypoints in point clouds  will recover annotated locations.
	
	Our approach is advantageous in the following aspects:
	(1) It provides novel views of a training image, which feeds unseen semantics and novel locations to the detector. 
	(2) The synthesized images are natural and have no artifacts. 
	(3)  Annotations are generated automatically for synthesized images; it requires no human labeling. 
	(4) We can control the number of novel images to generate for each existing training sample, and have the full flexibility in controlling the camera pose that corresponds to the rendered image. This ensures diversity and is a great advantage over traditional GAN-based methods, where 3D attributes of a new scene cannot be explicitly controlled.
	(5) Our proposed method is fully compatible with any kind of online data augmentation methods, \emph{e.g.}, affine transform, image mixup, \emph{etc.} Thus, the combined models will further improve the performance.
	
	In summary, we propose a cost-free approach to enrich images and labels, and show that it can significantly boost the performance of object detection systems suffering from scarce training data. 
	Contributions of this work include: 
	
	\bitem
	\item To the best of our knowledge, this is the first work to utilize the neural rendering for data augmentation in the object detection task. Our novel approach automates the annotation generation by rendering both images and annotations from cloud points.
	\item Our proposed method is 3D-aware and can augment long-tailed datasets in a controllable way. This method works for generic object detection and is compatible with online augmentation methods. 
	\item Extensive experiments show that the proposed method can significantly boost the performance of object detection systems; 
	the scarcity of training data is more severe, 
	the improvement achieved by our method is larger.
	\eitem

	\section{Related Work}
	\textit{Convolutional Neural Networks} (CNNs) have been successfully applied to computer vision tasks such as image classification \cite{he2019bag}, object detection \cite{zhou2019objects}, keypoint estimation \cite{sun2019deep}, image segmentation \cite{kirillov2020pointrend} and visual tracking \cite{wang2019fast}.
	However, CNN Models with inadequate generalizability may overfit the training data and perform poorly on unseen data.
	Improving the generalization ability of CNN models is a most demanding challenge to solve, especially when annotated data is scarce, unbalanced (with skewed ratios of majority to minority samples), or semantically impoverished.
	
	Existing approaches include semi-supervised learning methods \cite{chen2020big, zhai2019s4l, zoph2020rethinking} that interpolates labeled data from unlabeled data, self-supervised learning methods \cite{he2020momentum, chen2020simple} that exploit signals within unlabeled data via pretext tasks.
	Transfer Learning \cite{zhuang2019comprehensive}, One-shot and Zero-shot learning \cite{wang2019survey, wang2020generalizing} are other interesting paradigms for building models with extremely limited data. Conceptually similar to transfer learning, pre-training is training the network architecture on a big dataset such as ImageNet \cite{deng2009imagenet} or JFT-300M \cite{kolesnikov2019big}.
	In this paper, we focus on data augmentation which tackles the root of the problem, the training dataset, without changing the standard supervised learning paradigm.
	
	\subsection{Data Augmentation}
	Data Augmentation \cite{shorten2019survey} is a powerful method to mitigate the problem of data scarcity, as augmented data will represent a potentially more comprehensive set of data points, closing the gap between the training and testing sets.
	Usually, the methods can be classified as data warping and oversampling. 
	Data warping augmentations transform existing images while preserving the labels. Oversampling augmentations create synthetic instances to add into the training set, often used to re-sample imbalanced class distributions.
	Methods can also be categorized into online and offline depending on when the augmentation process occur.
	
	\noindent\textbf{Online Methods:}
	The earliest Data Augmentations started from simple transformations such as horizontal flipping, color space augmentations, and random cropping, followed by affine transformations, image mixup \cite{he2019bag, zhang2019bag}, random erasing \cite{zhong2020random}, feature space augmentation \cite{devries2017dataset}, etc.
	Applying meta-learning \cite{vanschoren2018meta} concepts from\textit{ Neural Architecture Search} (NAS) \cite{liu2018darts} to data augmentation has become increasingly popular with works such as Neural Augmentation \cite{perez2017effectiveness}, Smart Augmentation \cite{lemley2017smart}, and AutoAugment \cite{cubuk2018autoaugment}, respectively.
	Optimal online augmentation policies can be automatically searched within a search space. A policy consists of various sub-policies, one of which is chosen for each image in a mini-batch. 
	
	\noindent\textbf{Offline Methods:}
	Following the introduction of GANs \cite{goodfellow2014generative}, data augmentation itself is augmented with methods such as adversarial training \cite{goodfellow2014explaining}, GAN-based augmentation \cite{bowles2018gan}, neural style transfer \cite{gatys2015neural}.
	Being computationally expensive, GAN-based methods usually offer offline augmentation, \emph{i.e.}, constructing an extended dataset.
	Methods such as GANs and neural style transfer can 'imagine' alterations to images. However, such hallucination is artificial (often generated from random noise or limited conditional signals), the distribution of which may deviate from the original image, potentially leading the training process astray. 
	
	\begin{figure*}
		\centering
		\includegraphics[width=0.8\linewidth]{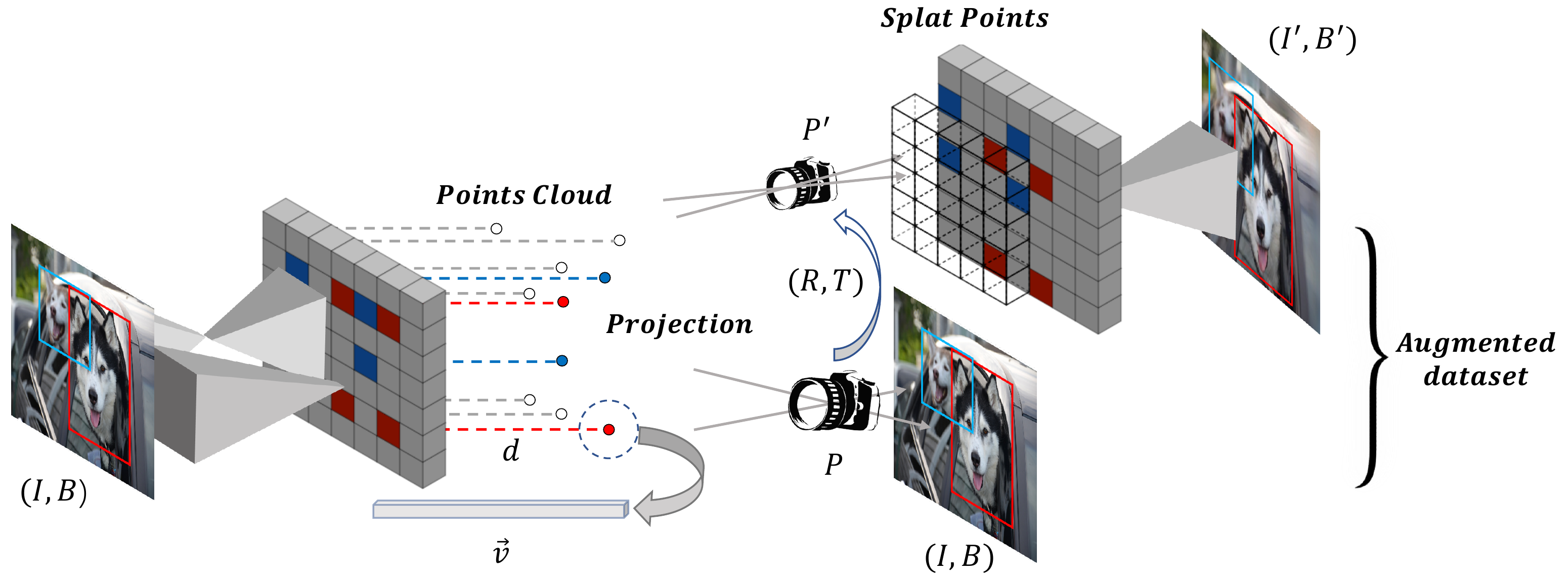}
		\caption{Neural renderer for object detection.} 
		\label{fig:renderer}
		\vspace{-5pt}
	\end{figure*}
	
	\subsection{Neural Rendering}
	In cognitive computer vision domain, most existing tasks are related to perception, \emph{i.e.}, perceive information from the image, video, voxel, mesh, or 3D point cloud. This is a process of  either 2D or 3D reasoning. Typical tasks include object detection, human pose estimation, segmentation, 3D shape estimation, etc.
	In computer graphics domain, rendering is the process of generating images of 3D scenes defined by geometry, materials, scene lights and camera properties.
	The purpose of neural rendering \cite{kato2020differentiable, tewari2020state} is to bridge the gap between the 2D and 3D processing methods, allowing neural networks to optimize 3D entities while operating on 2D projections.
	Applications of differentiable neural rendering include: novel view synthesis \cite{shih20203d, martinbrualla2020nerfw, Shih3DP20, wiles2020synsin}, semantic photo manipulation \cite{brock2016neural}, facial and body reenactment \cite{wu2020unsupervised, liu2019neural}, re-lighting \cite{chen2020neural}, free-viewpoint video \cite{martin2018lookingood}, creating photo-realistic avatars \cite{shysheya2019textured} or simply generating high-quality images \cite{gao2020sketchycoco}.
	Our proposed augmentation method is inspired by novel-view synthesis \cite{wiles2020synsin}, which is fully compatible with online data augmentation methods, and can be incorporated together to further inflate the dataset with novel semantics.
	GAN is described in \cite{bowles2018gan} as a way to ``unlock'' additional information from a dataset.
	With neural rendering, we further unlock dataset information in highly controllable ways. The interpolation of data is non-linear and offers novel spatial semantics in 3D, which is extremely valuable for object detection tasks.
	
	\section{Proposed Method}
	In this section, we introduce the proposed Dataset Augmentation system based on Neural Rendering. We denote our method as \textit{DANR}. 
	
	\subsection{Overview}
	The overview of the DANR system is illustrated in Fig. \ref{fig:overview}. The purpose of the system is to augment an object detection dataset with novel view images to improve the performance of object detectors. The amount of augment images and degree of camera pose variations are both controllable.
	At its core, the system builds upon a novel view synthesis model, as shown in Fig. \ref{fig:renderer}.
	The model takes as input an RGB image $I$ and a series of 2D image keypoints $B_{i}$ (representing bounding box annotations).
	The RGB image is simultaneously fed to two branches: (1) an hourglass network \cite{newell2016stacked} for depth estimation, and (2) an Encoder with inception-resnet blocks \cite{8064661} to extract visual features. 
	The visual features are pixel-aligned, \emph{i.e.}, each pixel on the feature maps carries a latent-space vector that corresponds to this particular spatial location. The depth estimation branch assigns depth value to each pixel, elevating latent-space vectors to a point cloud. Each point in the cloud resides in a 3D space with the 2.5D coordinate representation \cite{iqbal2018hand}:  $(u, v, d)$. Each point, however, may unfold its hidden N-dimensional latent-space vector after being projected to a target camera view.
	These feature vectors can be re-projected to an arbitrary camera coordinate system, based on a given projection matrix (translation vector $T$ and rotation vector $R$ for the corresponding camera of the original image).
	Given a target camera pose $P'$, we splat the feature points following Wiles's method \cite{wiles2020synsin} to make the rendering process differentiable.
	After splatted on a palette of size $K \times K$, these points result in an N-dimentional feature map of our chosen resolution. The feature maps are then fed to a Decoder network with inception-residual blocks and upsampling layers, rendering the final novel view image.
	
	In order to recover the bounding box annotations for object detection datasets, we mark the keypoints of the bounding boxes beforehand. 
	When they are projected to 3D space with depth information and re-projected back to 2D in a novel camera view, the marks consistently follow corresponding pixel-aligned points.
	The whole system is end-to-end differentiable.
	
	\subsection{DANR}
	In this section, we describe the data augmentation system in details. Firstly, we elaborate the state-of-the-art networks we employ and the specific settings. Then we introduce the point-cloud projection process and how to automate the generation of annotations in target views. Lastly, we depict the losses used to train the whole system.
	
	\noindent\textbf{Networks.}
	We employ hourglass network as our depth estimator, as this UNet-like structure is proved advantageous in terms of exploring global interdependencies across multiple scales. Estimating the depth of a pixel requires understanding of the global features as well as the local features, which is important in perceiving relative depth.
	We stack two level-4 hourglass networks, with maximum channel number $256$. We use a $1 \times 1$ filter at the last convolutional layer to produce the depth map.
	For the feature extractor network and refinement network after point splatting, we follow \cite{wiles2020synsin} to use Encoder-Decoder networks but makes several modifications: (1) when the input resolution is set to $512 \times 512$, we reduce the feature channel to $64$; (2) we replace the basic resnet block with inception-resnet block, as the concatenation of features further consolidates the representative power. 
	
	\noindent\textbf{Point Projection.}
	We fix the intrinsic parameters for all images, assuming they are taken with a default camera. Specifically, we set focal lengths on both $x$ and $y$ axes to be $1$ and set skew to be $0$, while assuming principal point is at the origin of coordinates. 
	Given a 2.5D point representation, we map it to the Camera Coordinate System (CCS) of the input image:
	\begin{equation}
		\label{eqn:pixel_to_camera}
		\left\{
		\begin{aligned}
			X_{c} = d \times (u - c_{x})  /  f_{x},  \\
			Y_{c} = d \times (v - c_{y})  /  f_{y},  \\
			Z_{c} = d. 
		\end{aligned}
		\right.
	\end{equation}
	
	This CCS is now regarded as the World Coordinate System (WCS). 
	Given a specific target view, we can determine the extrinsic parameters. Transforming from the WCS to a novel CCS is straight-forward with $R$ and $T$:
	\begin{equation}
		\label{eqn:world_to_camera}
		P_{c} = R \cdot ( P_{w} - T )
	\end{equation}
	where $P_{c} = (X_{c}, Y_{c}, Z_{c})$.
	
	The points $P_{c}$ are then splatted to a target camera palette with new pixel locations:
	\begin{equation}
		\label{eqn:camera_to_pixel}
		\begin{aligned}
			u' = f_{x} \times  X_{c} / Z_{c} + c_{x}, \\
			v' = f_{y} \times  Y_{c} / Z_{c} + c_{y}. \\
		\end{aligned}
	\end{equation}
	
	Within each bounding box $i$, points (at least the four points at corners) are marked with an index value $i$.
	These marked keypoints are a subset of $S \times S$ points, where $S$ is the resolution of the extracted feature maps.  
	When keypoints from the $i_{th}$ bounding box are re-projected to the target view after the procedures described above, their new pixel locations are denoted as $P'_{i} = \{(u'_{i}, v'_{i}) \}$. 
	The annotation for bounding box $i$ in the target image is computed as:
	
	\begin{equation}
		\label{eqn:new_annotation}
		\left\{
		\begin{aligned}
			x_{min} = \min(\{u \mid (u', v') \in P'_{i} \} ),\\
			x_{max} = \max(\{u \mid (u', v') \in P'_{i} \} ),\\
			y_{min} = \min(\{v \mid (u', v') \in P'_{i} \} ),\\
			y_{max} = \max(\{v \mid (u', v') \in P'_{i} \} ).
		\end{aligned}
		\right.
	\end{equation}
	
	\noindent\textbf{Loss Function.}
	We follow \cite{wang2018high} for the discriminator design, and use two multi-layer discriminators, the one with lower resolution serving as intermediate supervision to help the one with higher resolution learn better and faster.
	The overall loss is consisted of image-level L1 loss, feature-level content loss, and discriminator loss.
	The depth map is implicitly learned where we do not enforce supervised learning. The overall loss is:
	\begin{equation}
		\mathcal{L} = \lambda_{1} \mathcal{L}_{D}  + \lambda_{2} \mathcal{L}_{L1} + \lambda_{3} \mathcal{L}_{C}.
	\end{equation}

	\subsection{Keypoint-based Object Detection}
	Keypoint estimation is naturally a regression problem \cite{8064661}. The targets are represented by a series of heatmaps, each channel corresponding to a specific keypoint genre.
	Recent object detection methods such as CenterNet \cite{zhou2019objects} , CornerNet \cite{law2018cornernet} and ExtremeNet \cite{zhou2019bottom}, begin to utilize keypoint estimation techniques.
	For instance, CenterNet transforms the task of object detection from generating and classifying proposals into predicting objects' centers (keypoints) and corresponding attributes. To object detection, the attributes are the width, height of the object, along with local offsets that recover pixel location in the original resolution from down-sampled heatmaps. 
	
	We use CenterNet as our baseline detection framework to conduct experiments and ablation studies, due to its simplicity.
	Because it is anchor-free and does not need NMS as a post-processing step, the ablation study can be decoupled from complex design choices that is not concerned with training data.
	\noindent\textbf{Anchor-free.}
	During training, an object does not need to be assigned with proper anchor(s). Instead, only heatmaps for the entire image are generated as the regression target.
	\noindent\textbf{NMS-free}.
	When the heatmaps for object centers are inferred, the local peaks are ranked based on their response, and only top $K$ objects are extracted. With the center positions, corresponding attributes are extracted across their respective channels at the same 2D position. 
	Since keypoints of different genres can occur at the same position, CenterNet is also naturally compatible with multi-label problems.
	
	\subsection{Keypoint-based Mixup}
\begin{figure}
		\centering
		\includegraphics[width=0.95\linewidth]{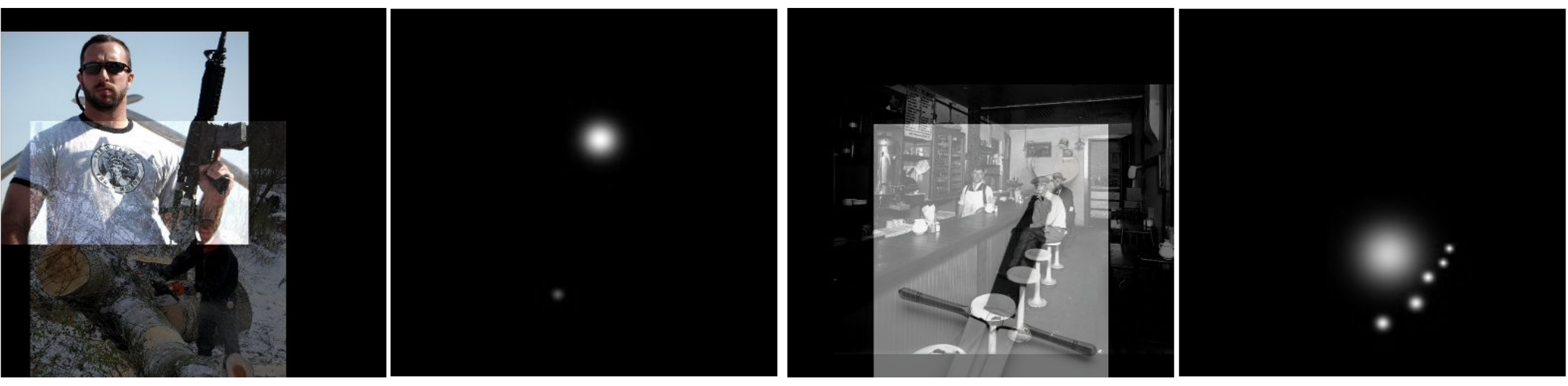}
		\caption{\textbf{Image mixup for keypoint-based object detection.} The radius of a heatmap blob is proportional to the size of the object, as illustrated in Equation \ref{eqn:radius}. Objects with higher weights correspond to keypoints with stronger response.} 
		\label{Fig:mixup}
		\vspace{-1em}
	\end{figure}

	It is first demonstrated in \cite{inoue2018data} how mixing samples up could be developed into an effective augmentation strategy.
	Image mixup for the task of object detection has been described in \cite{zhang2019bag}, but it is restricted to bounding box mixup. 
	We propose a  straight-forward image mixup strategy for keypoint-based object detection methods, where ground truth keypoints are splat onto a heatmap ${Y \in [0,1]^{\frac{W}{R} \times \frac{H}{R} \times C}}$ using a Gaussian kernel:
	\begin{equation}
		\label{eqn:radius}
		{Y_{xyc} = \exp\left(-\frac{(x-\tilde p_x)^2+(y-\tilde p_y)^2}{2\sigma_p^2}\right)}
	\end{equation}
	$\sigma_p$ is an adaptive standard deviation~\cite{law2018cornernet} that is proportional to the size of the object.
	During mixup, the confidence of keypoint heatmaps are applied with the same weights used on its corresponding image, as shown in Fig. \ref{Fig:mixup}. 
	This mixup strategy can be applied to keypoint-based object detection methods such as CenterNet.
	Compatible with the proposed offline augmentation method, image mixup is used in all of our experiments as an online augmentation method.
	
	\section{Experiments}
	
	\subsection{Dataset}
	
	\begin{table*}
		\centering
		\footnotesize
		\caption{Datasets used for neural rendering and object detection.}
		\label{tab:datasets}
		\begin{tabular}{c|cccccccc}
			\toprule
			\textbf{Dataset} & \textbf{Type} & \textbf{Domain} & \textbf{Purpose} & \textbf{Annotation} & \textbf{\#Category} & \textbf{\#Boxes}\\
			\midrule
			RealEstate10K      & Video & Indoor & Novel View Synthesis & - & - & - \\
			\midrule
			Indoor-Objects     & Image & Indoor & Object Detection & Bounding Box & 7 & 4,595  \\
			COCO               & Image & Indoor \& Outdoor & Object Detection & Bounding Box & 80 & 860,001  \\
			\midrule
			ICR-Weapon         & Image & Real-world Online Data & Object Detection & Bounding Box & 29 & 3,887 \\ 
			ICR-Flag           & Image & Real-world Online Data & Object Detection & Bounding Box & 1 & 2,002 \\ 
			ICR-Logo           & Image & Real-world Online Data & Object Detection & Bounding Box & 1 & 1,365  \\
			\midrule
			ICR-Norm10K        & Image & Real-world Online Data & Evaluation & - & - & -  \\
			\bottomrule
		\end{tabular}
	\end{table*}
	
	\noindent\textbf{Dataset: Data Augmentation.}
	RealEstate10K \cite{realestate10k} is a large dataset of camera parameters (intrinsic and extrinsic) corresponding to $10$ million frames derived from about $80,000$ video clips, gathered from about $10,000$ YouTube videos. For each clip, the poses form a trajectory where each pose specifies the camera position and orientation along the trajectory. These poses are derived by running SLAM and bundle adjustment algorithms on a large set of videos.
	We download Youtube videos with the given URLs and then extract image frames. 
	Note that we only use $10,758$ out of $82,315$ clips gathered from Youtube videos. 
	We also trained the neural renderer with Matterport3D \cite{Matterport3D} and Replica\cite{replica19arxiv}, synthetic datasets in the format of Habitat \cite{habitat19iccv}. However, we notice the model trained with Realestate10K has better generality as it is trained with real-world data instead of synthetic data. 
	In the following experiments, we use numerous augmentation systems that are trained on Realestate10K to aid the training of object detection networks.
	
	\noindent\textbf{Dataset: Object Detection.}
	To validate the effectiveness of DANR, we generate augmented datasets for Indoor object-detection \cite{indoor-dataset} dataset to mimic the constrained environment that is similar to RealEstate10K where we train the neural renderer.  
	We conduct experiments on Indoor dataset where we constrain to a subset of the dataset to simulate limited data problems.
	We also test on real-world data in an image content review system, which consists virtually any kind of images that users may upload. We organize these data into ICR-Weapon, ICR-Flag and ICR-Logo datasets. 
	Although not captured in a constrained environment, ICR benefits drastically from the proposed DANR method because ICR datasets suffer from limited positive samples in contrast to massive negative samples, where we need to improve the recall of positive samples while maintaining a very high through rate.
	We further attempt to push boundaries on the generalized object detection dataset COCO \cite{lin2014microsoft}.
	COCO is neither long-tailed nor insufficient in samples, so we perform DANR on COCO to measure its scope of limitations. 
	
	\subsection{Metrics}
	
	\noindent\textbf{Metrics: Synthesis Quality.}
	We use metrics peak-signal-to-noise ratio (PSNR) and structural similarity index measure (SSIM) to assess the objective image quality. 
	Since the metrics of  PSNR and SSIM are linked \cite{hore2010image}, we also resort to perceptual similarity (PSIM) \cite{zhang2018unreasonable} metric to compare the similarity of images.
	Although it remains challenging to determine the similarity of images when human judgement is involved \cite{zhang2018unreasonable}, these metrics altogether provide a more robust estimation of relative quality of synthesized images.
	We extract feature stack from $L$ layers and unit-normalize in the channel dimension, which we designate as $\hat{y}^l, \hat{y}_0^l \in \mathds{R}^{H_l\times W_l\times C_l}$ for layer $l$. We scale the activations channel-wise by vector $w^l \in \mathds{R}^{C_l}$ and compute the $\ell_2$ distance. Finally, we average spatially and sum channel-wise. Note that using $w_l=1 \forall l$ is equivalent to computing cosine distance.
	\vspace{-5pt}\begin{equation}
		d(x,x_0) = \sum_l \dfrac{1}{H_l W_l} \sum_{h,w} || w_l \odot ( \hat{y}_{hw}^l - \hat{y}_{0hw}^l ) ||_2^2
		\label{eqn:dist}
		\vspace{-5pt}
	\end{equation}
	
	\noindent\textbf{Metrics: Object Detection.}
	We use Average Precision (AP) as object detection metrics.
	For real-world image review datasets: ICR-Weapon, ICR-Flag, ICR-Logo, we report AP as detection metric and AUC  as review metric.

	\subsection{Implementation Details}
	For DANR, we crop input images to $512 \times 512$ for feature extraction and depth estimation. We also use a palette of the same size to splat points.
	We train on $4$ V100 GPUs with batch size $8$, randomly choosing samples for $500$ iterations per epoch. 
	The sampled viewpoints of a reference video frame and its paired video frame are within $30$ frame range.
	The network is trained for a total of $275$ epochs, which takes $3.5$ days.
	Here are the corresponding metrics of the model, evaluated on RealEstate10K validation set: (1) the SSIM metric is $0.4386$; (2) PSNR is $7.1564$ and (3) perceptual score is $0.5964$.
	
	\begin{figure}
		\centering
		\includegraphics[width=1.0\linewidth]{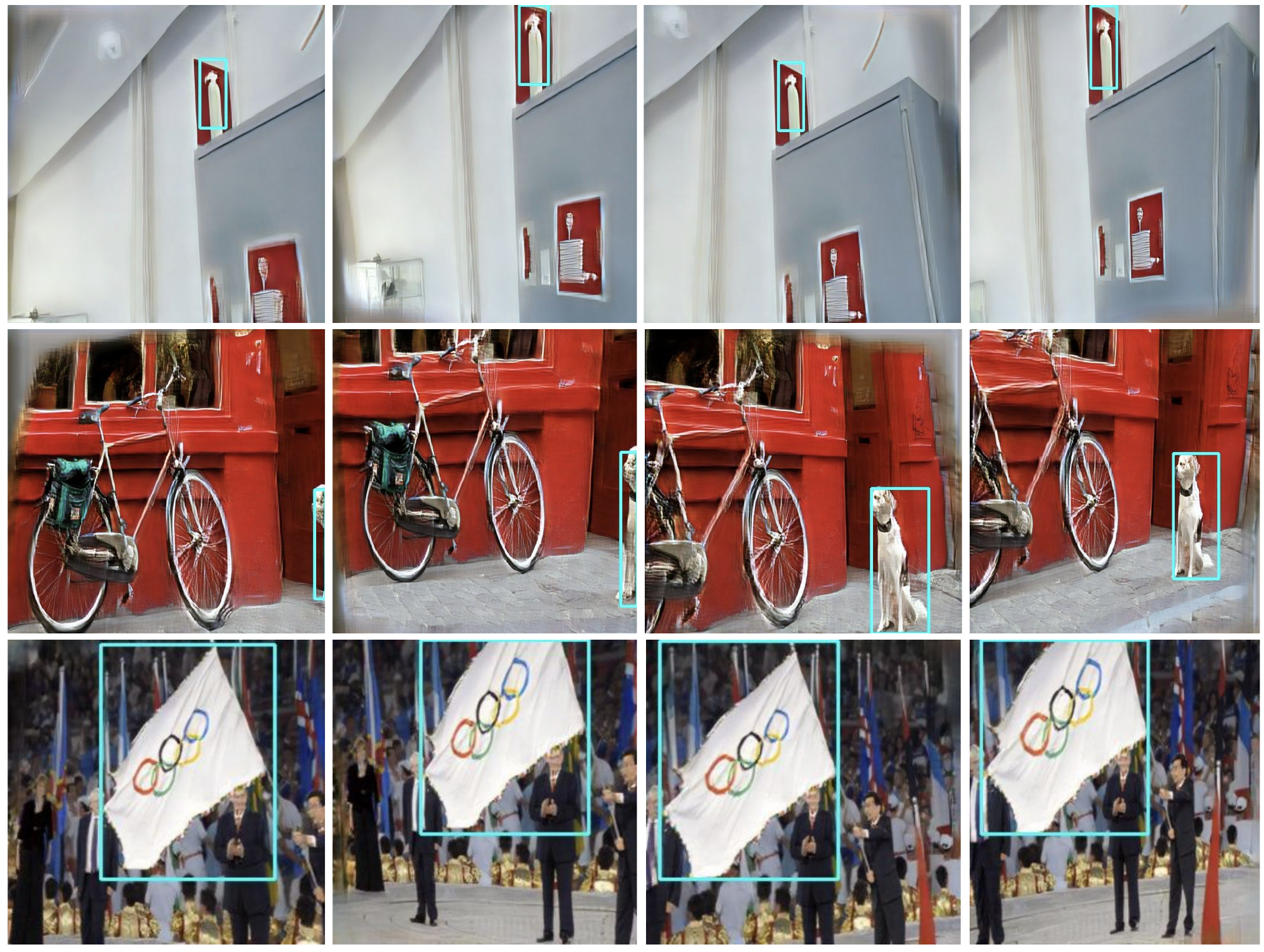}
		\caption{\textbf{Auto-generated annotations.} The bounding box annotations are precise for rendered images.} 
		\label{fig:anno_aug}
		\vspace{-1em}
	\end{figure}
	
	\subsection{Baseline}
	We use CenterNet as our baseline detection framework to conduct experiments and ablation studies.
	We use identical training settings for all ablation experiments.
	Regardless of backbones, the network performs affine transform to keep aspect ratio and use $512 \times 512$ image patch as input.
	Each network is trained for a total of $120$ epochs.
	We use Adam optimizer \cite{kingma2014adam}. Learning rate is initiated to $2.5 \times 10^{-4}$, which drops half twice at epoch $90$ and $110$, respectively.
	We do not perform any test-time augmentations, such as mirror flipping and multi-scale testing.
	For datasets except COCO, each image is augmented with four pre-defined views. They share the same translation vector 
	$\begin{bmatrix}
		0 & 0 & 0.3\\
	\end{bmatrix}$. 
	Their respective rotation vectors are: 
	$\begin{bmatrix}
		-0.1 & -0.15 & 0\\
	\end{bmatrix}$,
	$\begin{bmatrix}
		0.1 & -0.15 & 0\\
	\end{bmatrix}$,
	$\begin{bmatrix}
		-0.1 & 0.15 & 0\\
	\end{bmatrix}$,
	$\begin{bmatrix}
		0.1 & 0.15 & 0\\
	\end{bmatrix}$.
	For COCO dataset, we augment each image by randomly choosing one of the pre-defined views.

	\subsection{Ablation Study}
	In this section, we ablate various choices in the use cases of our proposed \textit{DANR}. For simplicity, all accuracy results here are for Indoor-Object validation set.
	Specifically, we show that the improvement is consistent regardless of detector configurations on various data distributions. We also discuss additional characteristics of augmentation such as the impact of image resolution. The following comparisons will be made.
	(1) Augmentation: online vs. offline; 
	(2) render: high-resolution vs low-resolution;
	(3) Affine Transform: w/ affine vs. w/o affine;
	(4) data scarcity level: low  vs. high;
	(5) detector: keypoints-based (one-stage) vs proposal-based (two-stage);
	(6) backbone: heavy vs. lightweight.
	
	\noindent\textbf{Compatibility:} 
	We check the compatibility of the proposed DANR with online augmentation methods. In our baseline setting, we use CenterNet as our detection framework, with a pre-trained ResNeSt-50 backbone. 
	We add a keypoint-based image mixup strategy to represent online augmentation, denoted as \textbf{\textit{Online}}. 
	We add the data generated offline by the DANR to training, denoted as \textbf{\textit{Offline}}. 
	In all of the following experiments, we only compare \textit{+Online} vs. \textit{+Online+Offline}, denoted as \textit{N.A}. and \textit{+Aug}, respectively.
	\begin{table}[t]
		\centering
		\begin{tabular}{l c@{\ \ }c @{\ \ } c@{\ \ }c @{\ \ } c@{\ \ }c }
			\hline
			Aug-types &  AP & $AP_{50}$ & $AP_{75}$ & $AP_{S}$ & $AP_{M}$ & $AP_{L}$ \\
			\hline
			Baseline & 67.2 & 94.7 & 77.5  & 58.8 & 60.9 & 70.5   \\
			\hline
			+ Online & 69.4 & 95.4 & 78.4  & 56.5 & 59.1 & 72.8   \\
			\qquad \quad \enspace \ + Offline & 77.8 & 96.9 & 90.4  & 54.8 & 59.9 & 81.5   \\
			+ Online + Offline  & 78.4 & 97.3 & 90.8  & 54.2 & 59.8 & 82.1   \\
			
			\hline
		\end{tabular}
		\caption{\textbf{Compatibility: Online vs. Offline. }Performance for augmentation types on Indoor validation set. }
		\label{tab:compatibility}
		\vspace{-2em}
	\end{table}
	Table \ref{tab:compatibility} shows that DANR is compatible with online mixup. Their augmentation strategies are complementary. DANR alone can boost the performance by around 10 percent.
	
	\noindent\textbf{Render Resolution:} 
	With DANR, the detection performance is significantly improved. However, we notice that the average precision for small objects suffer a performance drop. We hypothesize that it is due to the small render resolution (\emph{e.g.} $256 \times 256$) compared to the detector input resolution (\emph{e.g.} $512 \times 512$).
	
	\begin{figure*}
		\centering
		\includegraphics[width=0.9\linewidth]{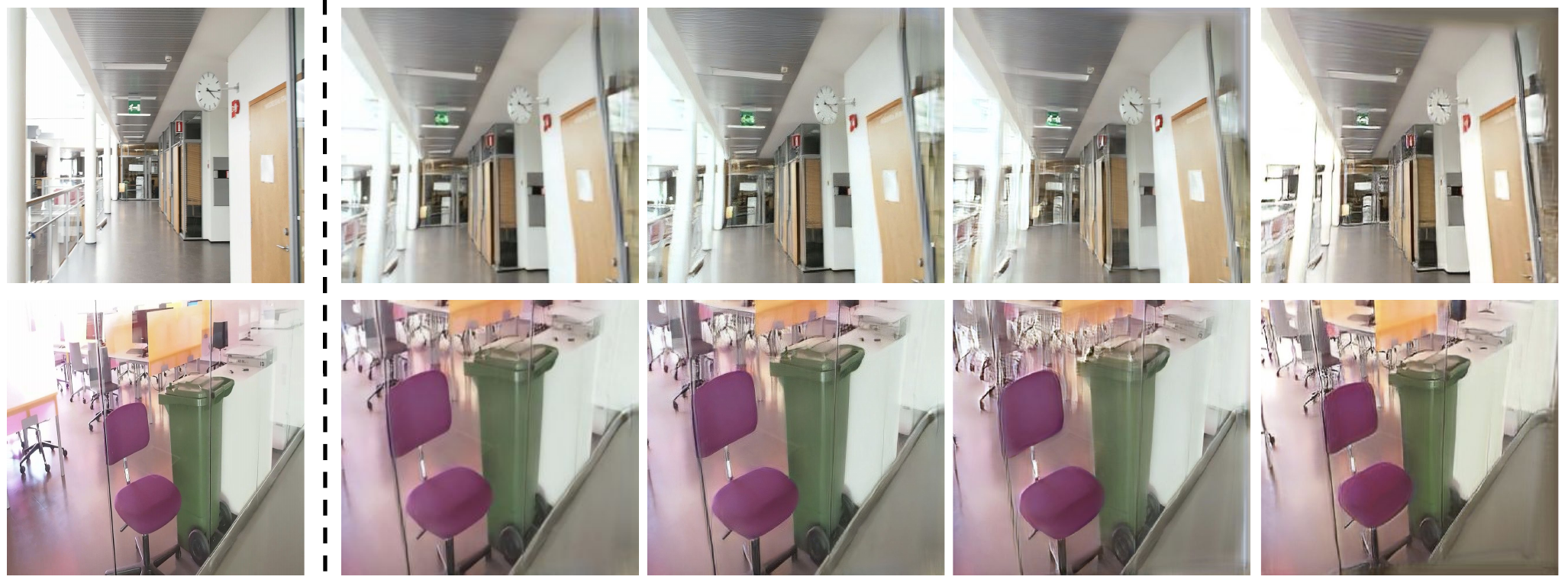}
		\caption{\textbf{Augmentation with different rendering resolutions and techniques for Indoor dataset}. Columns from left to right: Original image, rendered images with: 256-Native, 512-PixelSS, 512-PointSS, 512-Native.} 
		\label{fig:image_resolution}
		\vspace{-1em}
	\end{figure*}
	
	In order to generate images at higher resolution, we experiment with 3 different settings:
	(1) Without any further training, we apply the network trained on a smaller image size to a different size at test time.
	Specifically, we splat the $256^{2}$ cloud points onto $512 \times 512$ palette maps, then render the maps to an image with a refinement network.
	This augment method increased image resolution by super-sampling of splatted points, and is denoted as \textbf{\textit{512-pointSS}}.
	(2) We first augment images at $256 \times 256$, and further upsample pixels at the image level with a pre-trained super-sampling network \cite{wang2018esrgan}. 
	We denote this method as \textbf{\textit{512-pixelSS}}.
	(3) We re-train a network from scratch that takes in images at $512\times512$, generate feature maps at $512 \times 512$, and sample $512^{2}$ feature points in the cloud, which are splatted and rendered on the output image.
	This method naturally outputs images in $512\times512$ resolution, denoted as \textbf{\textit{512-native}}.
	
	\begin{table}[ht]
		\small
		\centering
		\begin{tabular}{l c@{\ \ }c @{\ \ } c@{\qquad}c @{\ \ } c@{\ \ }c }
			\hline
			Render-Res &  AP & $AP_{50}$ & $AP_{75}$ & $AP_{S}$ & $AP_{M}$ & $AP_{L}$ \\
			\hline
			N.A. & 69.4 & 95.4 & 78.4  & 56.5 & 59.1 & 72.8   \\
			\hline
			256-native & 78.4 & 97.3 & 90.8  & 54.2 & 59.8 & 82.1   \\
			512-pixelSS & 79.0 & 97.5 & 91.4  & 63.2 & 65.5 & 82.5   \\
			512-pointSS & 79.1 & 97.6 & 92.0  & 64.5 & 70.4 & 82.4   \\
			512-native & 79.6 & 97.7 & 91.4  & 61.4 & 67.3 & 83.2   \\
			\hline
		\end{tabular}
		\caption{\textbf{Render Resolution: High-Res vs. Low-Res. }
			Performance on Indoor validation set with augmentation at different resolutions. 
		}
		\label{tab:resolution}
		\vspace{-1em}
	\end{table}
	As illustrated in Table \ref{tab:resolution},
	(1) augmentation with images rendered at different resolutions consistently boosts the detection performance;
	(2) synthesized images at low resolutions may potentially lose some details compared to real images, which does harm to the detection performance of very small objects;
	(3) uplifting the image resolution via super-sampling further improves the performance;
	(4) super-sampling from points may be a better choice than super-sampling from pixels;
	(5) training a network with dense features and points achieves the best performance.
	In the following experiments, we use 512-native as our resolution setting, as it achieves the highest AP and does not need any additional super-sampling networks.
	
	\noindent\textbf{Affine Transform:} 
	As illustrated in Table~\ref{tab:affine},
	(1) depriving affine transform from the default online augmentation during training impairs the performance;
	(2) offline data augmentation improves performance with or without affine transformation; 
	(3) to some extent, offline augmentation can make up for the loss of affine transform during training.
	\begin{table}[t]
		\small
		\centering
		\begin{tabular}{l c@{\ \ }c @{\qquad} c@{\ \ }c @{\qquad} c@{\ \ }c }
			\hline
			&  \multicolumn{2}{c}{AP} & \multicolumn{2}{c}{$AP_{50}$}  & \multicolumn{2}{c}{$AP_{75}$}\\
			Baselines & N.A. & +Aug & N.A. & +Aug & N.A. & +Aug \\
			\hline
			w/ affine & 69.4 & 79.6 & 95.4  & 97.7 & 78.4 & 91.4 \\
			w/o affine  & 57.7 & 79.0 & 93.0  & 98.0 & 63.9 & 91.3   \\
			\hline
		\end{tabular}
		\caption{\textbf{Affine Transform: w/ affine vs. w/o affine.} }
		\label{tab:affine}
		\vspace{-1em}
	\end{table}
	
	\noindent\textbf{Data Scarcity:} 
	We further experiment with three data-split settings: 5:5, 3:7 and 1:9.
	Highest degree of scarcity is 1:9.
	While splitting images to train and validation sets, an image is assigned along with its corresponding augmented images.
	\begin{table}[ht]
		\centering
		\begin{tabular}{l c@{\ \ }c @{\qquad} c@{\ \ }c @{\qquad} c@{\ \ }c }
			\hline
			&  \multicolumn{2}{c}{AP} & \multicolumn{2}{c}{$AP_{50}$}  & \multicolumn{2}{c}{$AP_{75}$}\\
			Train/Val & N.A. & +Aug & N.A. & +Aug & N.A. & +Aug \\
			\hline
			5/5 & 69.4 & 79.6 & 95.4  & 97.7 & 78.4 & 91.4 \\
			3/7 & 56.7 & 71.0 & 87.7  & 91.9 & 65.3 & 83.1 \\
			1/9 & 11.9 & 47.2 & 27.5  & 74.8 & 8.5 & 51.4 \\
			\hline
		\end{tabular}
		\vspace{-0.5em}
		\caption{\textbf{Scarcity Level: High vs. Low. }Performance (AP) on Indoor validation set with different split settings. The train:val ratio is 5:5, 3:7, 1:9, respectively.}
		\label{tab:scarcity}
		\vspace{-1em}
	\end{table}
	Table \ref{tab:scarcity} shows that the performance boost is closely related with the scarcity of training data. When annotated data is very limited, the DANR approach becomes more beneficial.

	\noindent\textbf{Backbone:} 
	We fix the detector to CenterNet while comparing different backbones. 
	\begin{table}[t]
		\small
		\centering
		\begin{tabular}{l c@{\ \ }c @{\qquad} c@{\ \ }c @{\qquad} c@{\ \ }c }
			\hline
			&  \multicolumn{2}{c}{AP} & \multicolumn{2}{c}{$AP_{50}$}  & \multicolumn{2}{c}{$AP_{75}$}\\
			Backbone & N.A. & +Aug & N.A. & +Aug & N.A. & +Aug \\
			\hline
			ResNet-50 & 53.2 & 77.4 & 77.0  & 96.4 & 60.3 & 90.7   \\
			ResNeSt-50 & 69.4 & 79.6 & 95.4  & 97.7 & 78.4 & 91.4 \\
			\hline
		\end{tabular}
		\caption{\textbf{Backbone: ResNet-50 vs. ResNeSt-50.} }
		\label{tab:backbone}
		\vspace{-1em}
	\end{table}
	Table \ref{tab:backbone} shows that the proposed augmentation consistently boost the detection performance, regardless of the backbones used by the detector.
	
	\begin{table}[ht]
		\small
		\centering
		\begin{tabular}{l c@{\ \ }c @{\qquad} c@{\ \ }c @{\qquad} c@{\ \ }c }
			\hline
			&  \multicolumn{2}{c}{AP} & \multicolumn{2}{c}{$AP_{50}$}  & \multicolumn{2}{c}{$AP_{75}$}\\
			Framework & N.A. & +Aug & N.A. & +Aug & N.A. & +Aug \\
			\hline
			One-Stage & 69.4 & 79.6 & 95.4  & 97.7 & 78.4 & 91.4 \\
			Two-stage & 68.1 &  78.7 & 96.4 & 97.5 & 81.5 &  89.3 \\
			\hline
		\end{tabular}
		\caption{\textbf{Detector: One-stage vs. Two-stage.} Performance for different detectors on Indoor validation set. We show that augmentation is consistently helpful regardless of detection frameworks. We use CenterNet as an example of one-stage method, while using Hit-Detector as a two-stage method.}
		\label{tab:detector-stage}
		\vspace{-1em}
	\end{table}

	\begin{table*}[ht]
		\small
		\centering
		\begin{tabular}{l c@{\ \ }c @{\qquad} c@{\ \ }c @{\qquad} c@{\ \ }c @{\qquad} c@{\ \ }c @{\qquad} c@{\ \ }c @{\qquad} c@{\ \ }c @{\ \ } c@{\ \ }c}
			\hline
			& \multicolumn{2}{c}{AP} & \multicolumn{2}{c}{$AP_{50}$}  & \multicolumn{2}{c}{$AP_{75}$} & \multicolumn{2}{c}{$AP_{S}$}  & \multicolumn{2}{c}{$AP_{M}$} & \multicolumn{2}{c}{$AP_{L}$} & \multicolumn{2}{c}{$AUC$}\\
			Dataset & N.A. & +Aug & N.A. & +Aug & N.A. & +Aug & N.A. & +Aug & N.A. & +Aug & N.A. & +Aug & N.A. & +Aug \\
			\hline
			Indoor & 69.4 & 79.6 & 95.4 & 97.7 & 78.4 & 91.4 & 56.5 & 61.4 & 59.1 & 67.3 & 72.8 & 83.2 & - & -\\
			ICR-Weapon & - & - & -  & - & - & - & - & - & - & - & - & - & 0.68 & 0.75 \\
			ICR-Logo & - & - & -  & - & - & - & - & - & - & - & - & - & 0.73 & 0.96 \\
			ICR-Flag & - & - & -  & - & - & - & - & - & - & - & - & - & 0.72 & 0.97\\
			COCO & 33.4 & 33.1 & 52.8  & 52.6 & 34.8 & 34.5 & 12.8 & 12.6 & 38.0 & 37.6 & 51.6 & 51.8 & - & -\\
			\hline
		\end{tabular}
		\caption{Performance on different data settings. We use CenterNet as our framework, with ResNeSt-50 backbone. N.A. denotes settings without DANR (with online augmentation); +Aug denotes settings with DANR.}
		\label{tab:results}
		\vspace{-1em}
	\end{table*}
	
	\begin{figure*}[ht]
		\centering
		\includegraphics[width=0.9\linewidth]{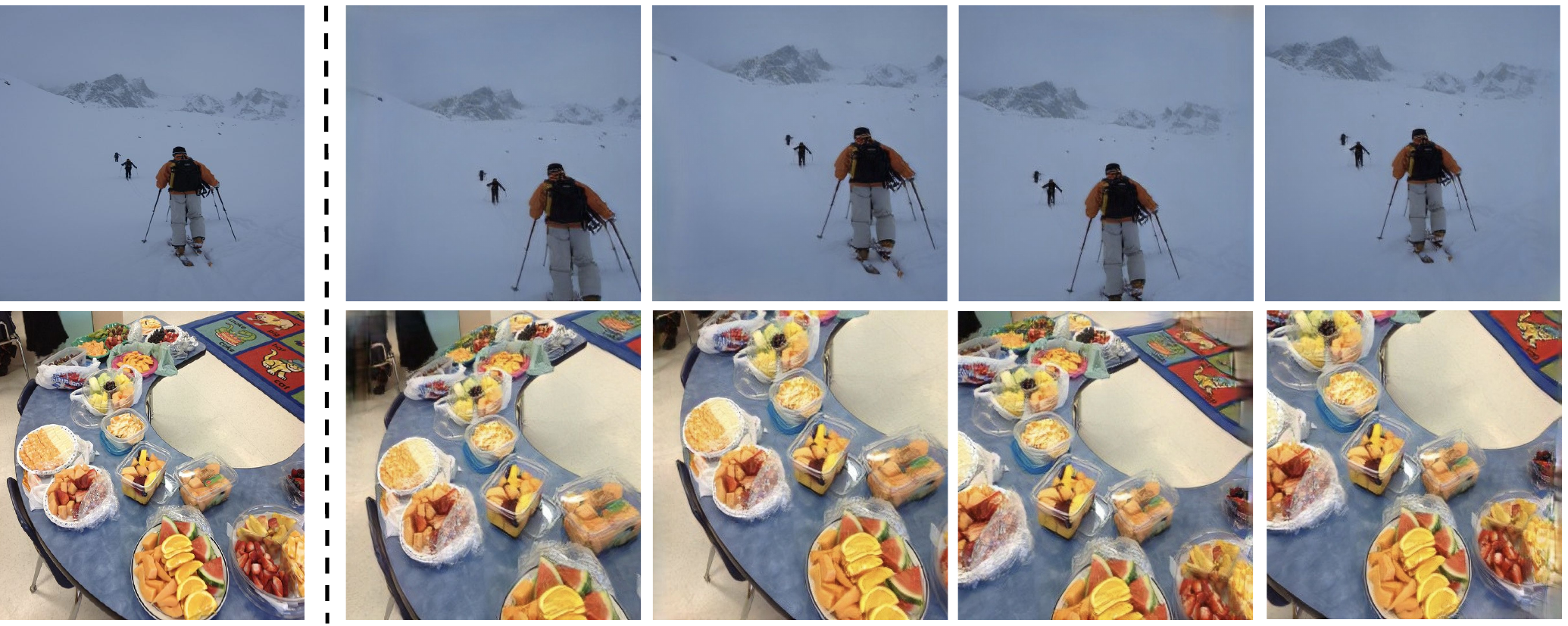}
		\caption{\textbf{Image augmentation results for COCO dataset.} Columns from left to right: Original image, rendered image views 1 to 4.
			The extreme case is, when depth for each pixel is identical, the augmentation is equivalent to affine transform, which is linear.
		} 
		\label{fig:image_aug}
		\vspace{-1.5em}
	\end{figure*}

	\noindent\textbf{Detector:} 
	Lastly, we compare one-stage and two-stage detection frameworks.
	We use CenterNet as an example of one-stage method, while using Hit-Detector \cite{guo2020hit} to represent the two-stage approach. We use the official code release of Hit-Detector to perform training and testing.
	Not surprisingly, \textit{DANR} does not rely on the one-stage detection framework. It is beneficial to two-stage detectors as well.
	
	\subsection{Experimental Results}
	
	Summarizing the experimental results on various datasets, as shown in Table~\ref{tab:results}, we have come to the following insights:
	(1) Results on image content review datasets (ICR-Weapon, ICR-Flag and ICR-Logo) show that the proposed DANR is consistently helpful on real-world data when training data is scarce.
	(2) Results on indoor dataset show that the augmentation is highly beneficial with indoor data, even when training data is not scarce. 
	(3) Results on COCO dataset have shown that DANR makes little contribution under circumstances where (a) the neural renderer in DANR fails to fully generalize to the semantics of the particular dataset, and (b) the training data is abundant.
	
	\section{Discussions}
	Overall, the role DANR plays in training an object detector on COCO dataset is neglectable, although the recall on large objects have improved over networks trained with the raw COCO dataset. 
	However, the overall AP is lower due to inferior performance on small objects. 
	Experimental results show that when annotated data is abundant,  COCO dataset for instance,  (1) augmented images with lower resolution may impair the network's ability in detecting small objects; (2)  the augmentation quality becomes crucial in boosting object detection performance, which is highly correlated with the synthesis resolution as well as the accuracy in depth estimation. 
	For models trained with indoor datasets, if augmenting COCO images in lower resolution, it does not provide COCO so much novel semantic data as to make up for its loss resulted from resolution degradation.  
	In short, DANR needs more outdoor training data to improve its generality. 
	This limitation will be mitigated with the maturity of future generative models.

	\section{Conclusion}
	In this paper, we have successfully developed a new method of data augmentation for object detection based on differentiable neural rendering. 
	Compared with existing methods, neural rendering offers a more promising generative modeling technique for use in data augmentation, as it bridges the 2D and 3D domain, and offers highly controllable ways to manipulate images. 
	Extensive experimental results show that the proposed method significantly improves the detection performance when annotated data is scarce, and is a very promising counterpart in dealing with long-tailed problems, along with other machine learning paradigms such as semi-supervised learning.
	In future research, we foresee the augmentation based on neural rendering effectively applied to other cognitive vision tasks.

	\newpage
	{\small
		\bibliographystyle{ieeetr}
		\bibliography{egbib}
	}
	
	\clearpage
	
	\appendix
	
	\twocolumn[{%
	\renewcommand\twocolumn[1][]{#1}%
	\begin{center}
		\bf \Large \modelname{}: Appendix
		\vspace{2em}
	\end{center}
	
}]
\renewcommand{\thesection}{\Alph{section}}

We give additional qualitative results in Section \ref{sec:results},
additional details of datasets in Section \ref{sec:datasets}, 
additional  architectural details in Section \ref{sec:arch},
and finally more information about the neural renderer in Section \ref{sec:differentiable}.

\section{Additional Qualitative Results}
\label{sec:results}
\begin{figure}
	\center
	\begin{overpic}[width=1\linewidth]{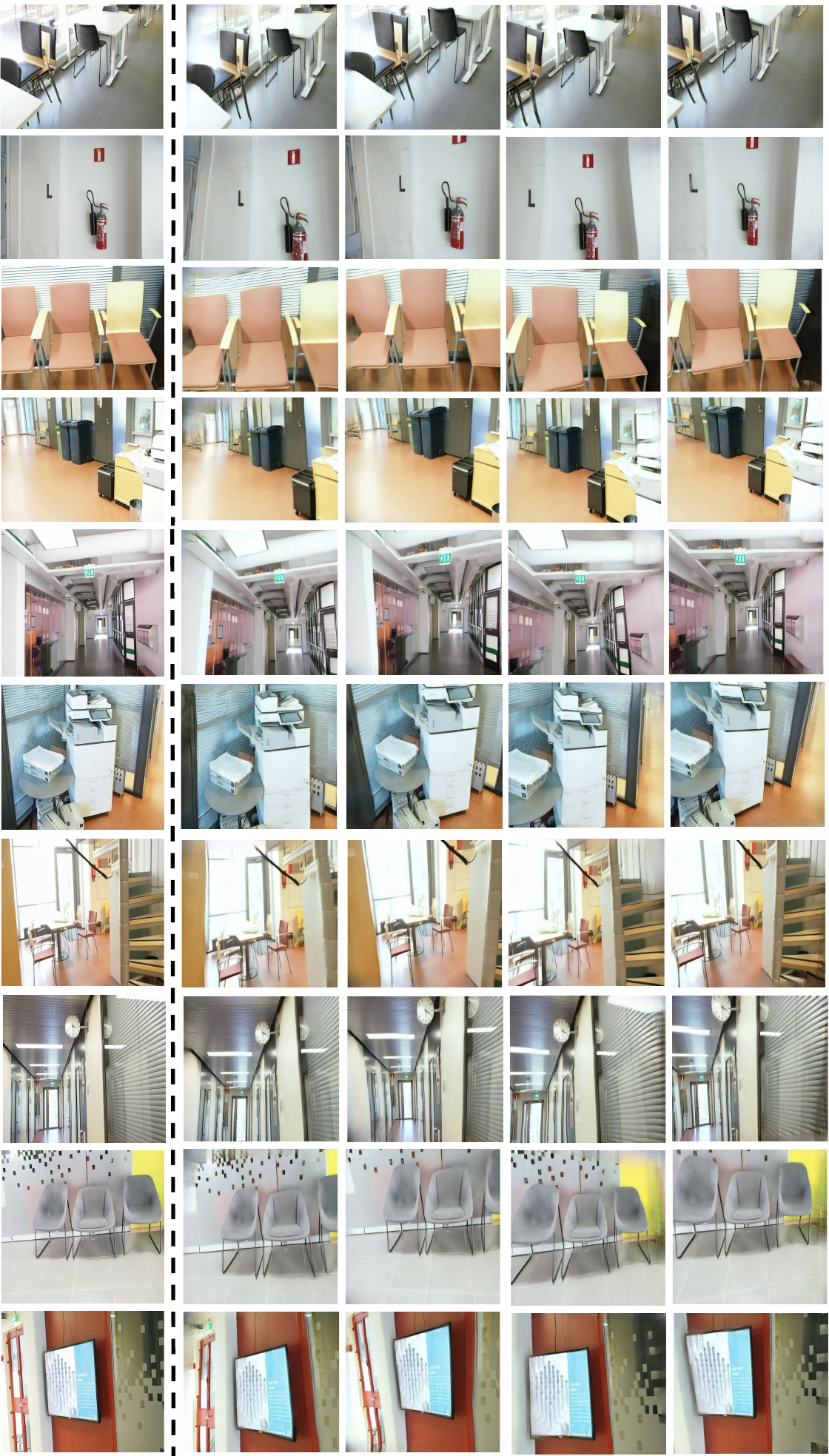}
		\scriptsize
		\put (1,-1.5) {Original Img}
		\put (30,-1.5) {Novel Views}
	\end{overpic}
	\vspace{2em}
	\caption{Additional results of augmented images on Indoor Objects. Zoom in for details.}
	\label{fig:Indoor}
\end{figure}

\begin{figure}
	\begin{overpic}[width=1\linewidth]{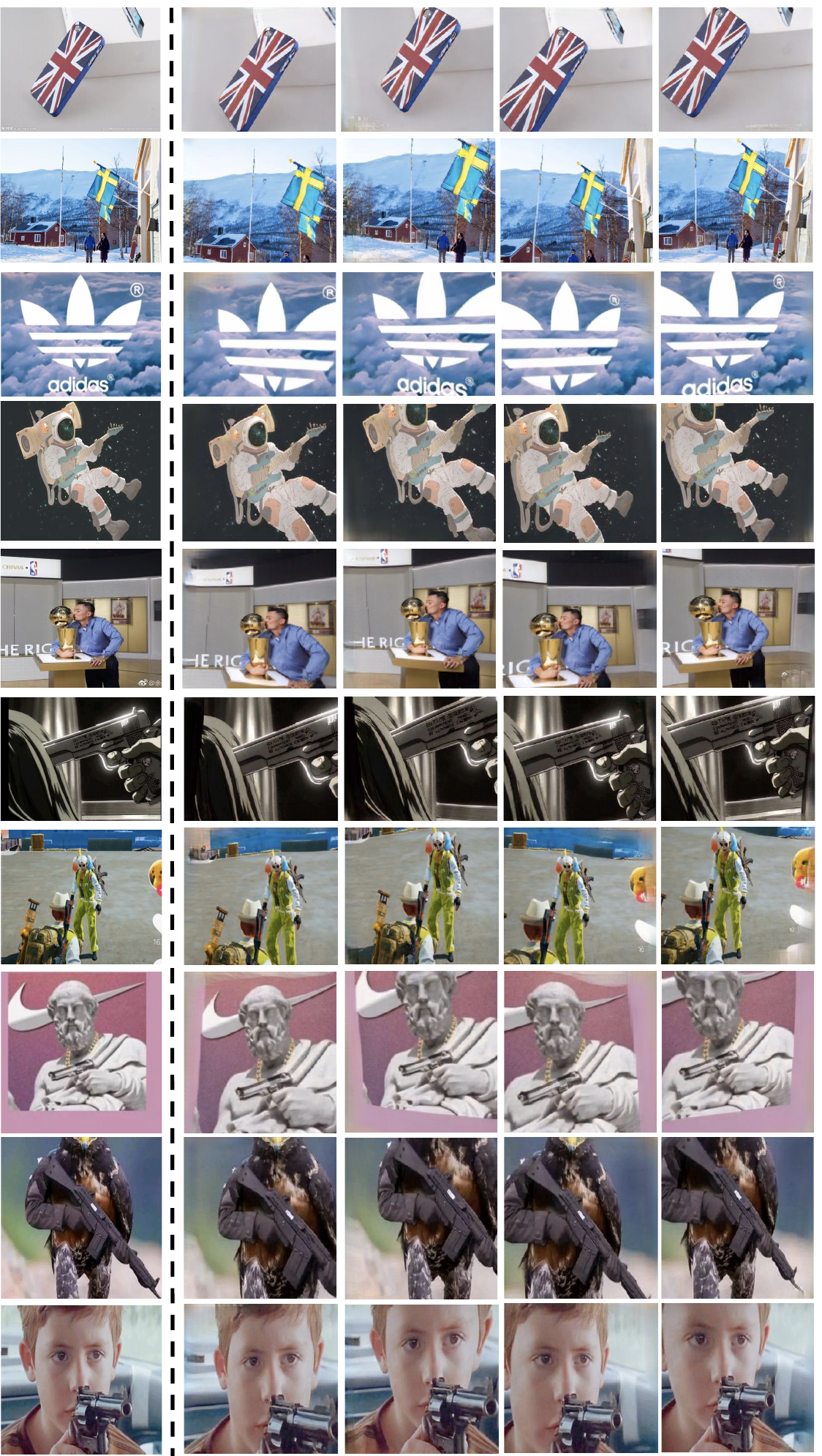}
		\scriptsize
		\put (1,-1.5) {Original Img}
		\put (30,-1.5) {Novel Views}
	\end{overpic}
	\vspace{1em}
	\caption{Additional results of augmented images on ICR-Flag, ICR-Logo and ICR-Weapon. The images are carefully selected due to the sensitivity of the data. Zoom in for details. }
	\label{fig:ICR}
\end{figure}

\begin{figure}
	\begin{overpic}[width=1\linewidth]{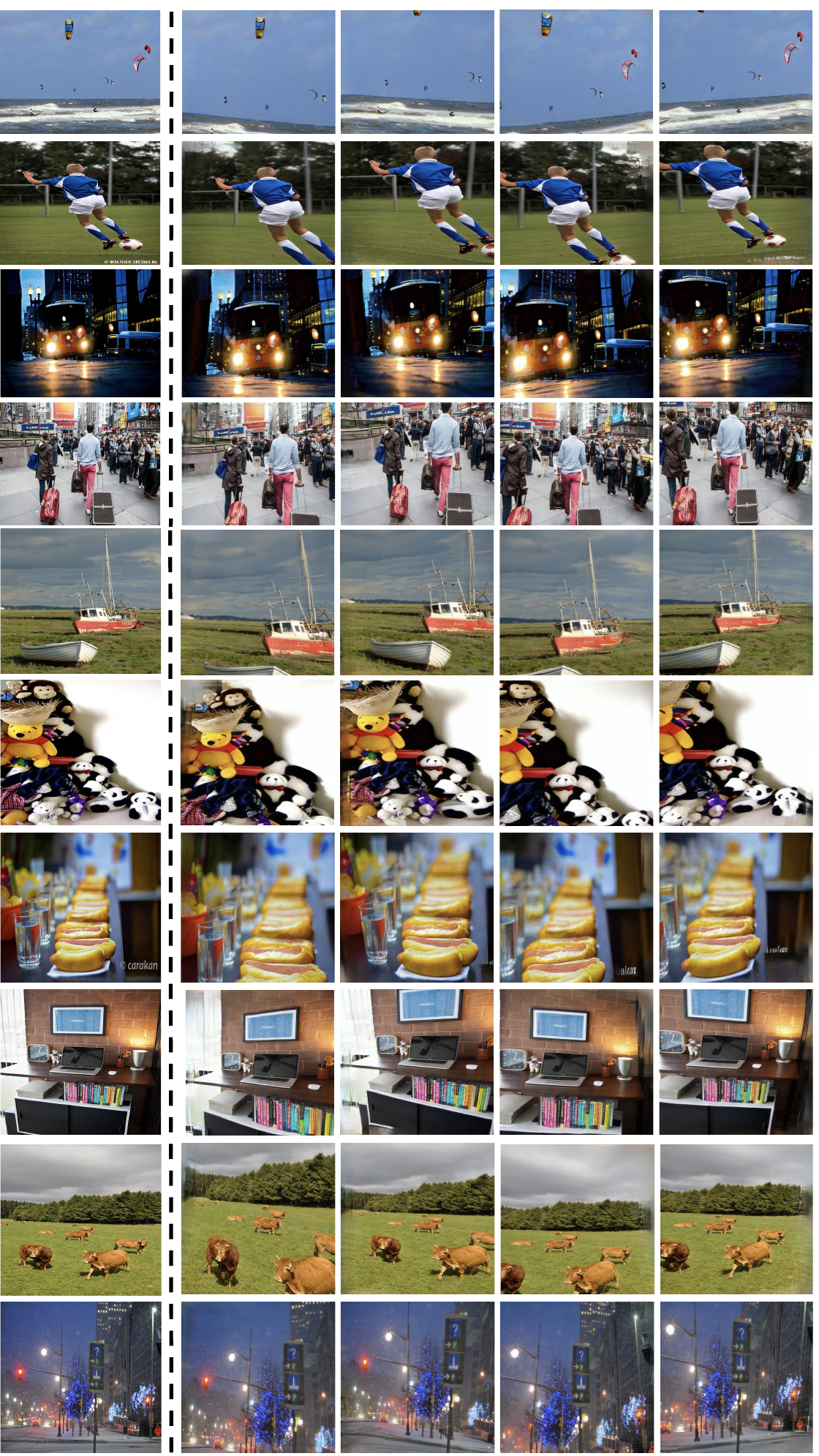}
		\scriptsize
		\put (1,-1.5) {Original Img}
		\put (30,-1.5) {Novel Views}
	\end{overpic}
	\vspace{2em}
	\caption{Additional results of augmented images on COCO. Zoom in for details.}
	\label{fig:COCO}
\end{figure}

\begin{figure}
	\begin{overpic}[width=1\linewidth]{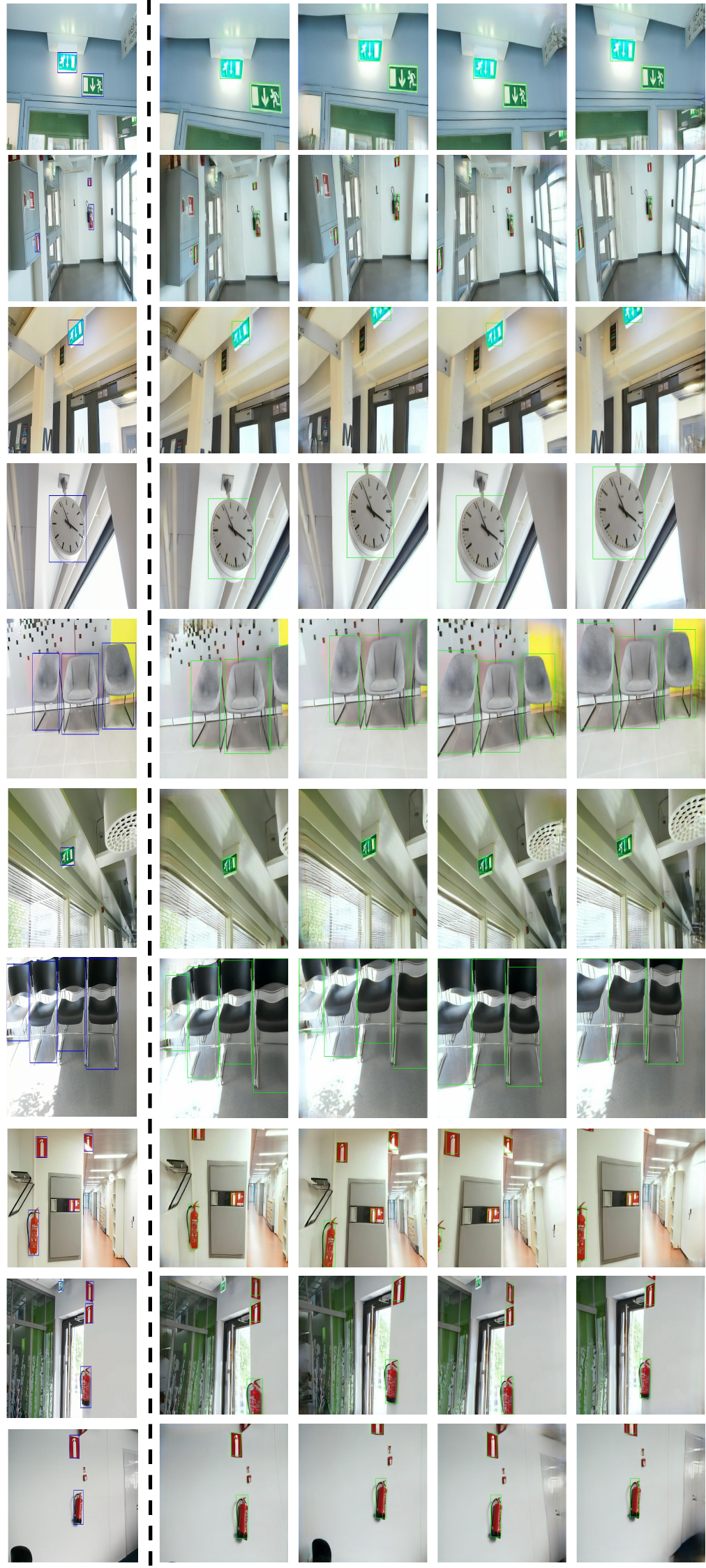}
		\scriptsize
		\put (5,-1.5) {GT }
		\put (20,-1.5) {Auto-Annotations for Novel Views}
	\end{overpic}
	\vspace{2em}
	\caption{Additional results of automatic annotations on Indoor dataset. Zoom in for details.}
	\label{fig:indoor_bbox}
\end{figure}

\begin{figure}
	\begin{overpic}[width=1\linewidth]{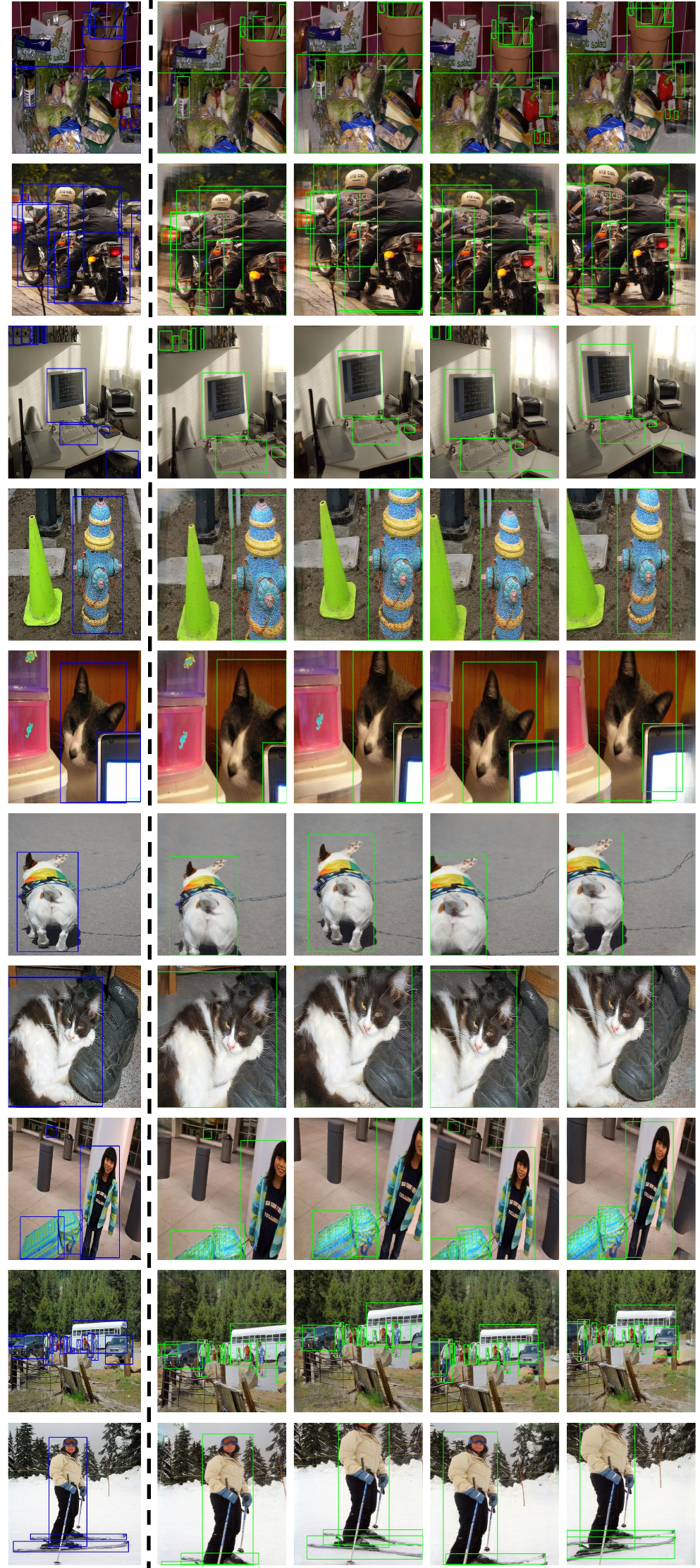}
		\scriptsize
		\put (5,-1.5) {GT }
		\put (20,-1.5) {Auto-Annotations for Novel Views}
	\end{overpic}
	\vspace{2em}
	\caption{Additional results of automatic annotations on COCO dataset. Zoom in for details.}
	\label{fig:COCO_bbox}
\end{figure}

We give additional qualitative results, including rendered images with novel views on Indoor (\figref{fig:Indoor}), ICR (\figref{fig:ICR}), and COCO (\figref{fig:COCO}). We also show automatically generated bounding box annotations (\figref{fig:indoor_bbox} and \figref{fig:COCO_bbox}).

\section{Additional Details on Datasets}
\label{sec:datasets}
\noindent\textbf{ICR dataset.} 
Collected from an online image content review system, the dataset is extremely unbalanced as positive samples are very rare compared to the negative samples, with a ratio of 1:100.
The positive samples are images potentially risky to exhibit in a system that seeks to retain certain types of visitors. The negative samples, on the other hand, are images safe to display.

One of the key metrics for evaluating such a review system is AUC score of the precision-recall curve. Specifically, the recall rate of positive samples and through rate of negative samples are measured for various thresholds.
It is easy to overfit to the limited number of positive samples and mis-classify large amounts of negative images, because the visual features of the positive samples can be prevalently found in the negative samples.
The ICR dataset is a much larger dataset compared to Indoor Objects dataset, in terms of the negative samples involved in training. 

Simply feeding images with empty heatmaps, our baseline CenterNet is advantageous in training with negatives samples. On top of the baseline, CenterNet with the proposed DANR is proven extremely helpful in improving the AUC score on ICR datasets, including ICR-Flag, ICR-Logo and ICR-Weapon.

\newpage
\section{Additional Details on Network Architecture}
\label{sec:arch}

Here we illustrate the precise details of network architectures used to build the components of DANR.

\paragraph{\bf Inception-Residual blocks.}

\begin{figure*}
	\centering
	\subfigure[][Inception-Residual block.]{\includegraphics[width=0.30\linewidth]{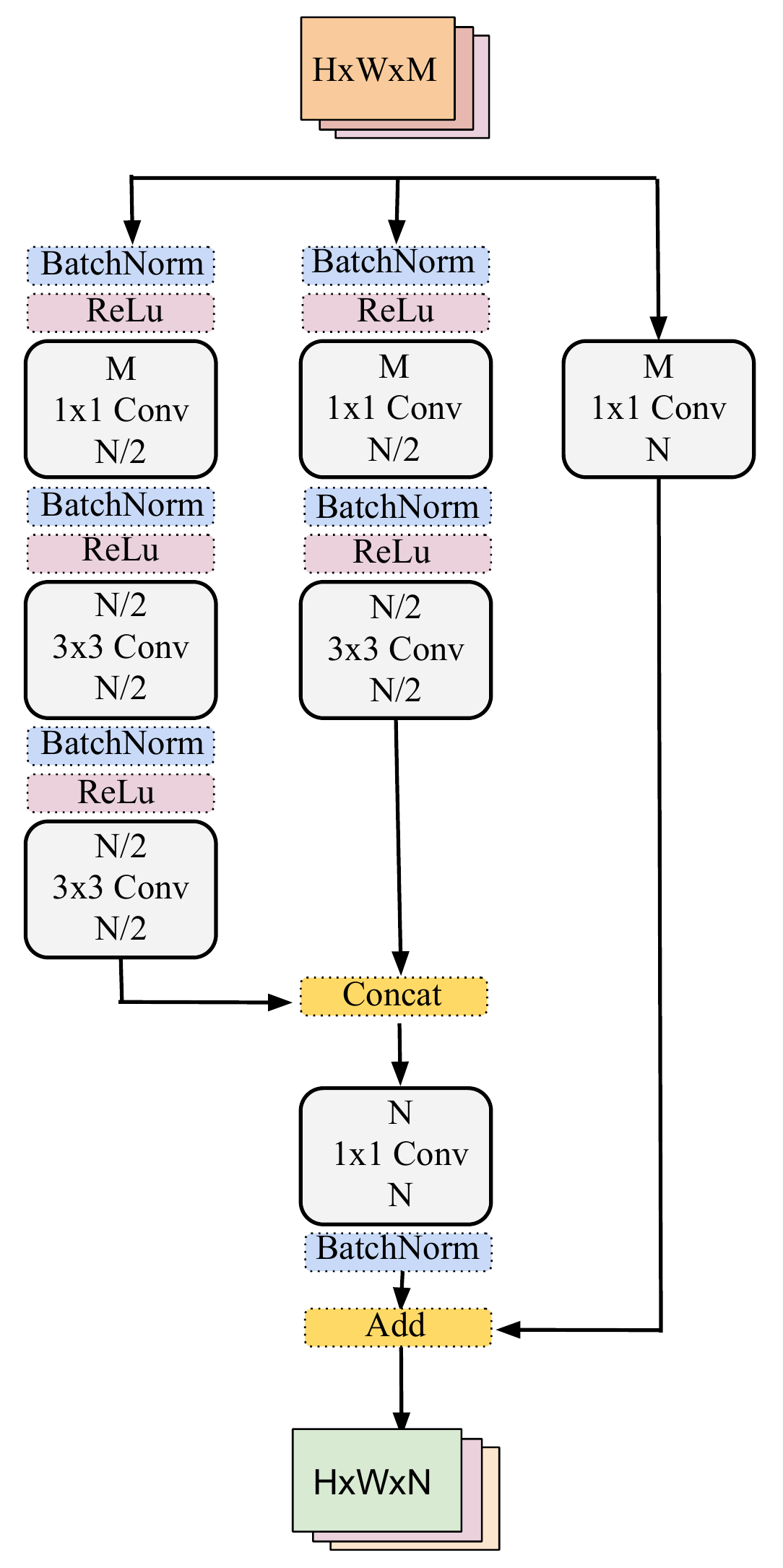}\label{fig:identity}} \quad \quad
	\subfigure[][Inception-Residual block with upsampling  layers.]{\includegraphics[width=0.30\linewidth]{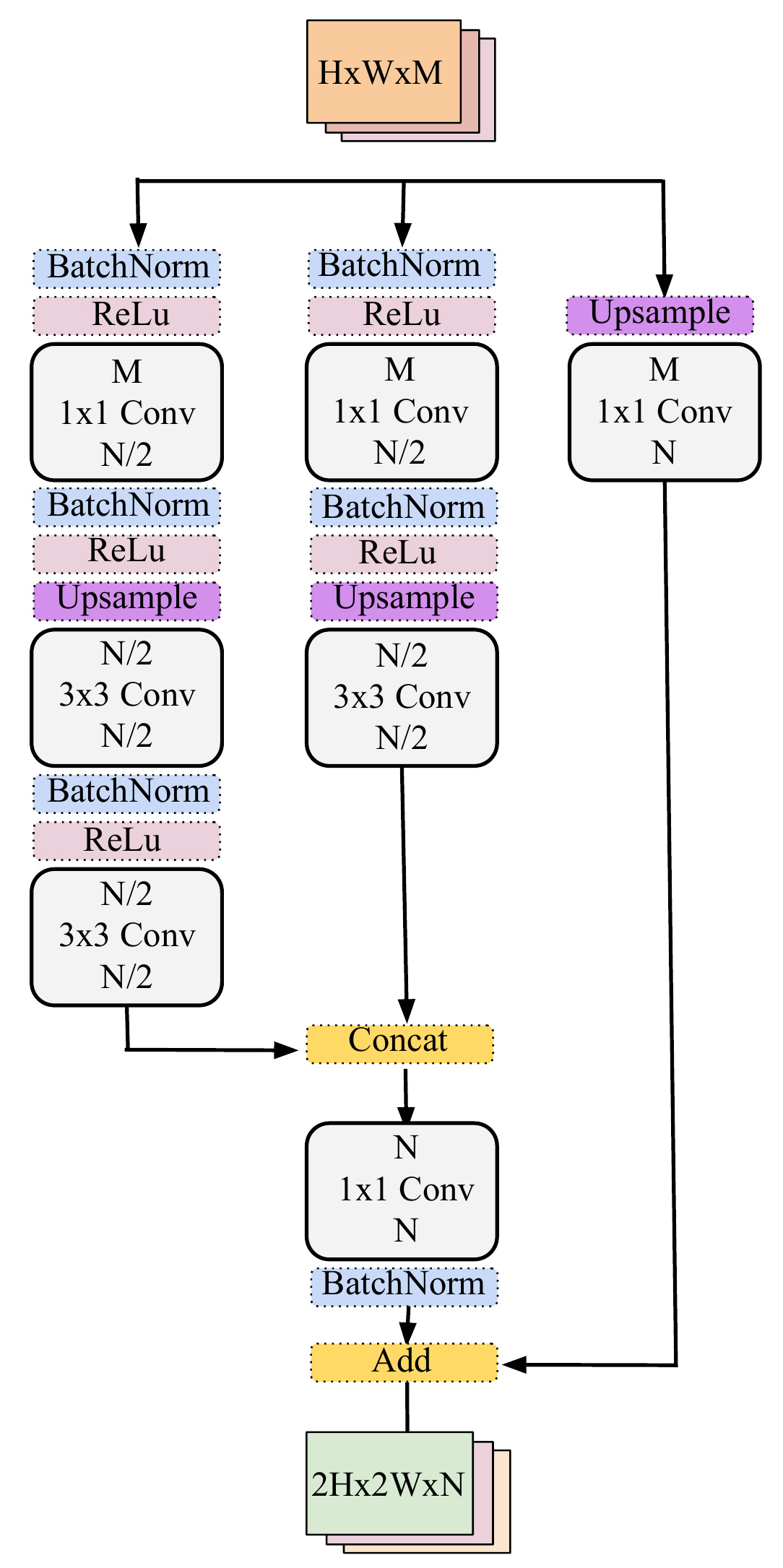}\label{fig:upsample}} \quad \quad
	\subfigure[][Inception-Residual block with average pooling layers.]{\includegraphics[width=0.30\linewidth]{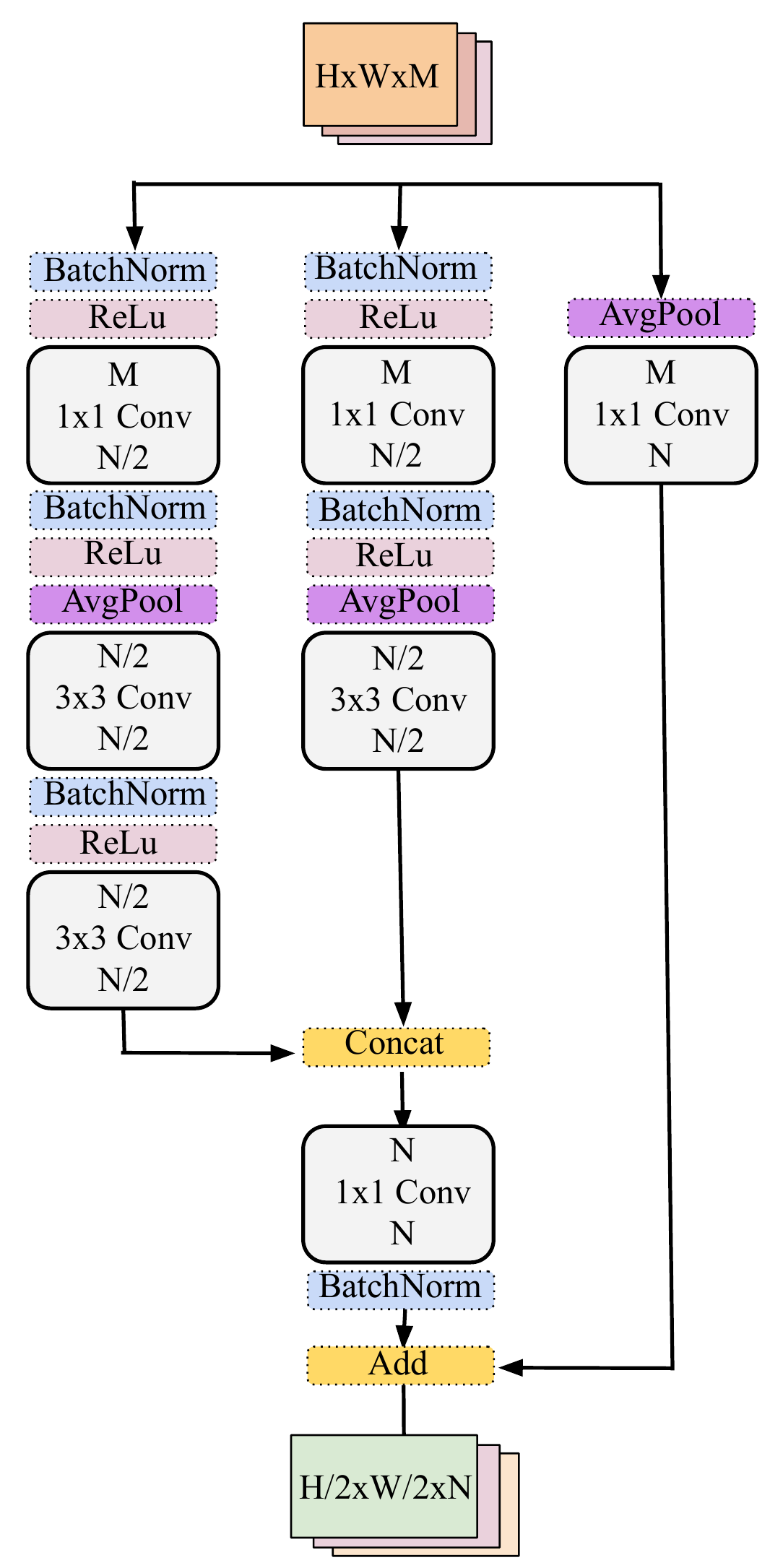}\label{fig:downsample}}
	\caption{An overview of Inception-Residual blocks. We show the basic block in (a) that maintains the resolution, in (b) when we uplift the resolution, and in (c) when we downscale the resolution.}
	\label{fig:inception-residual-block}
\end{figure*}

\begin{figure}
	\centering
	\subfigure[][Encoder.]{\includegraphics[height=0.8\linewidth]{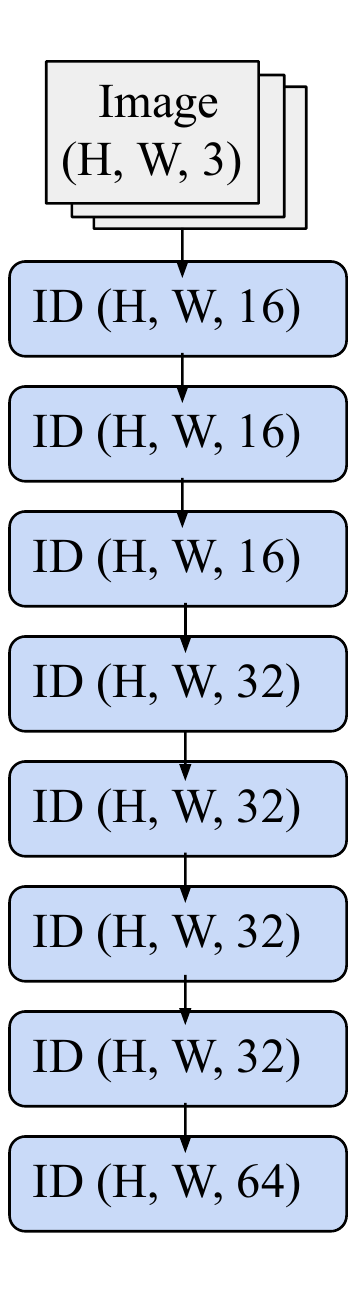}\label{fig:encoder}} \quad \quad
	\subfigure[][Decoder.]{\includegraphics[height=0.8\linewidth]{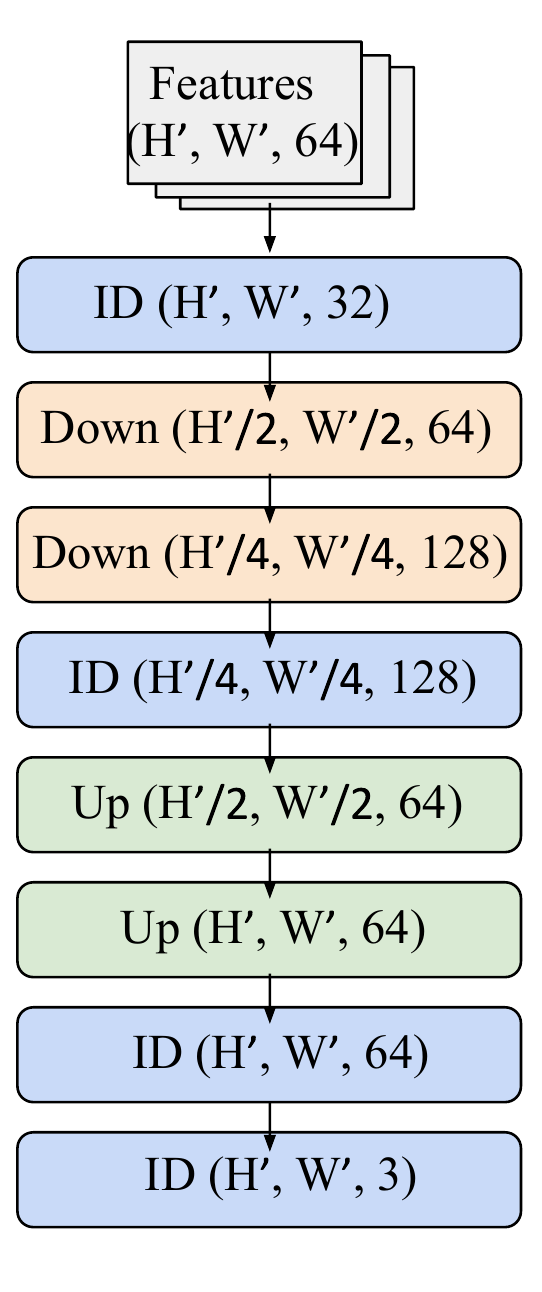}\label{fig:decoder}}
	\caption{Encoder-Decoder Networks built with aforementioned inception-residual blocks: Identity (ID), Upsampling (Up), Average Pooling (Down). The resolution of input features of the Decoder (H', W') is by default identical to the features extracted by the Encoder. When the 3D points are splatted to a larger palette, e.g., $\times2$ resolution, the feature map is upsampled by the scale of 2, which will uplift the image render resolution. We refer to this super-resolution method as \textit{\textbf{PointSS}}, in contrast to \textit{\textbf{PixelSS}} where the rendered image are fed to an external super-resolution network.}
\end{figure}

We first introduce the inception-residual block, the basic module that incorporates both inception design and ResNet design, i.e., channel-wise concatenation of two branches and pixel-wise addition with an identity branch.
As illustrated in \figref{fig:inception-residual-block}, the block can be configured to maintain, increase or decrease the resolution of features, with an identity layer (\figref{fig:identity}), an upsampling layer (\figref{fig:upsample}) and an average pooling layer (\figref{fig:downsample}, respectively.

\paragraph{\bf Encoder.}
Inception-residual blocks are stacked together to form the embedding network.
Specifically, our setup is illustrated in \figref{fig:encoder}.

\paragraph{\bf Decoder.}
Similarly, inception-residual blocks are stacked together to form the decoder network.
Our specific setup is shown in \figref{fig:decoder}.
The resolution of input features of the Decoder (H', W') is by default identical to the features extracted by the Encoder. When the 3D points are splatted to a larger palette, e.g., $\times2$ resolution, the feature map is upsampled by the scale of 2, which will uplift the image render resolution. We refer to this super-resolution method as \textit{\textbf{PointSS}}, in contrast to \textit{\textbf{PixelSS}} where the rendered image are fed to an external super-resolution network.

\begin{figure}
	\centering
	\includegraphics[height=1.0\linewidth]{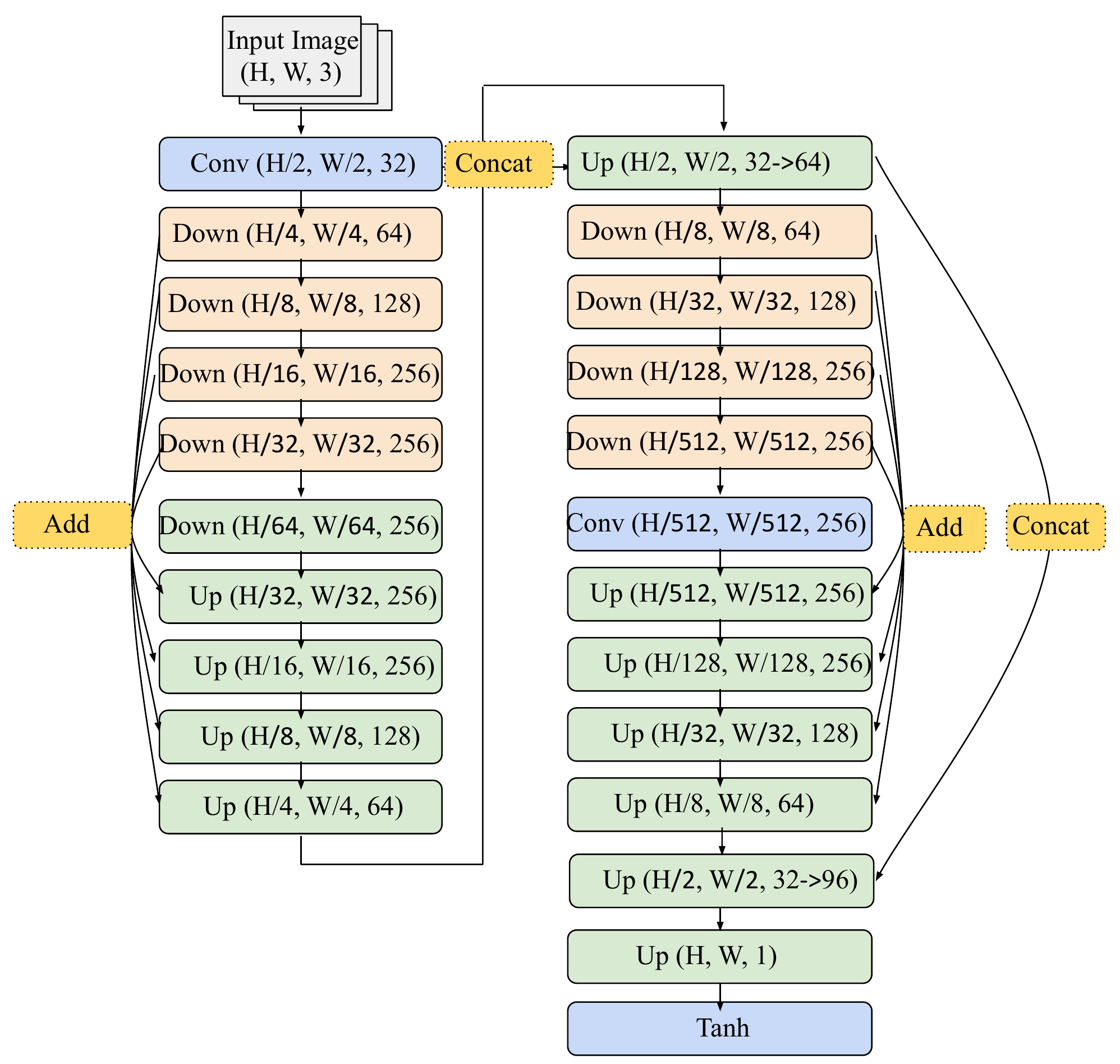}
	\caption{Depth estimation network. We simplify the hourglass network by replacing the \textit{pooling layer + resnet blocks} scheme with a {\em Down-sampling  Block} that consists of a sequence of Leaky ReLU, convolution (stride 2 or 4, padding 1, kernel size 4), and batch normalization layers.
		A {\em Up-sampling Block} consists of a sequence of ReLU, a 2x or 4x bilinear upsampling, convolution (stride 1, padding 1, kernel size 3), and batch normalization layers. We concatenate the output feature maps of stacked hourglass sub-networks and regress depth at the last layer (no batch normalization layer).}
	\label{fig:hourglass}
\end{figure}

\paragraph{\bf Depth estimator.}
The depth estimation network uses an hourglass architecture, as illustrated in \figref{fig:hourglass}.
We simplify the original hourglass network by replacing the \textit{pooling layer + resnet blocks} scheme with a {\em Down-sampling  Block} that consists of a sequence of Leaky ReLU, convolution (stride 2 or 4, padding 1, kernel size 4), and batch normalization layers.
A {\em Up-sampling Block} consists of a sequence of ReLU, a 2x or 4x bilinear upsampling, convolution (stride 1, padding 1, kernel size 3), and batch normalization layers. We concatenate the output feature maps of stacked hourglass sub-networks and regress depth at the last layer (no batch normalization layer).

\newpage
\section{Additional Details on Differentiable Rendering}
\label{sec:differentiable}

\begin{figure}
	\centering
	\includegraphics[width=0.5\linewidth]{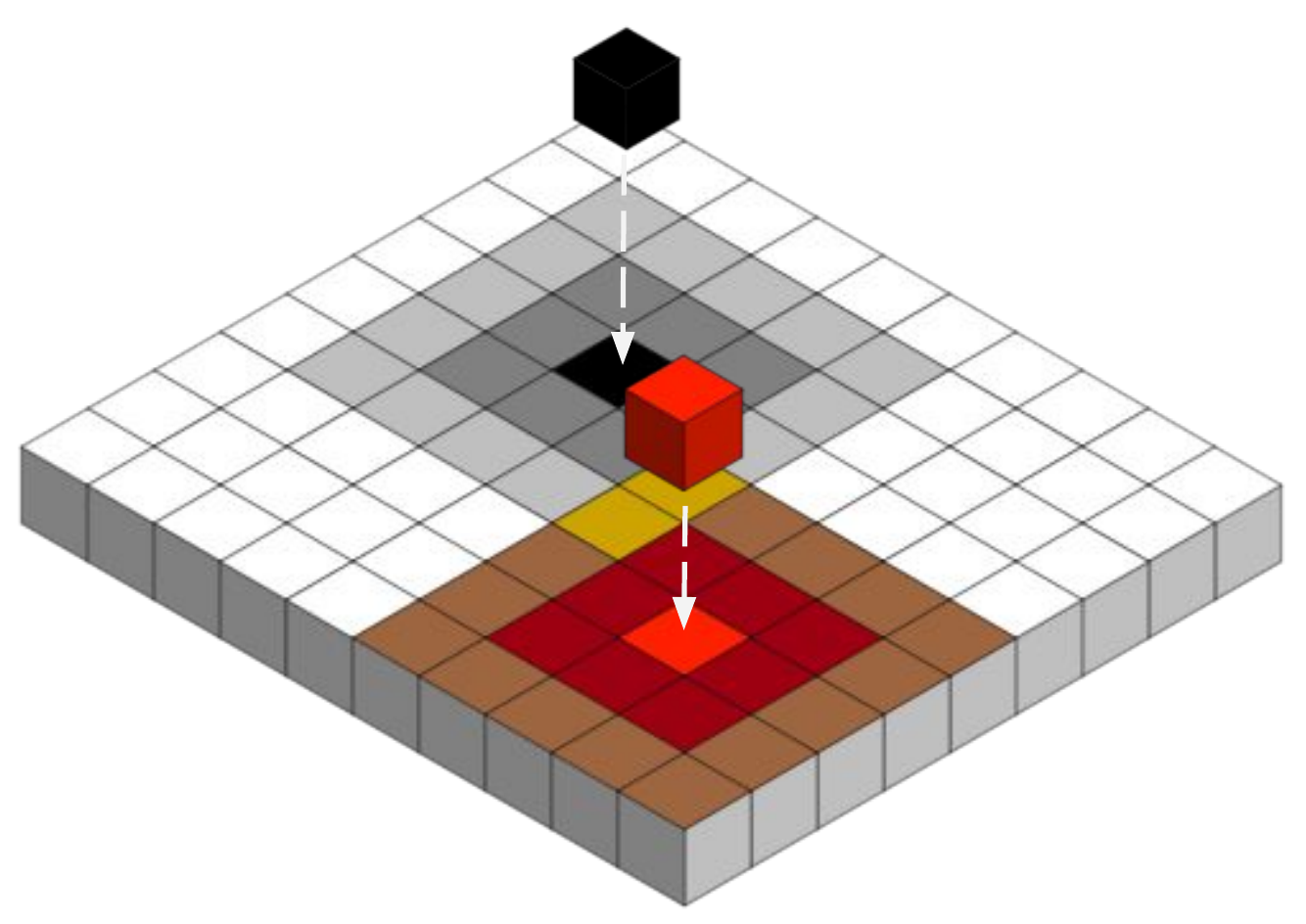}
	\caption{Differentiable renderer. In this example, two points from the z-buffer are splatted onto a palette with an influence range of radius $3$.}
	\label{fig:zbuffer}
\end{figure}

Before being splatted onto a palette, 3D cloud points in the new view are sorted in depth using a z-buffer.
Naively, for all points in the new view, the nearest point in depth
(by poping the z-buffer) is chosen to color that point.
We follow the differentiable renderer in order to provide gradients with respect to the point cloud positions.
In the differentiable renderer, $K$ nearest points for each pixel are splatted; each of these points influences a range that originates from the splatted pixel with radius $r$, the influence of which is proportional to the Euclidean distance from the center. 
The sorted points are then accumulated using linear alpha over-compositing.
In this way, the hard z-buffer becomes differentiable.
An example is shown in \figref{fig:zbuffer}.
In our usage case of DANR, $K = 128$, $r = 4$.

\end{document}